\documentclass[journal]{IEEEtran}
\usepackage{cite}
\usepackage{amsmath,amssymb,amsfonts}
\usepackage{algorithmic}
\usepackage{graphicx,color}
\usepackage{textcomp}
\usepackage{xcolor}
\usepackage{hyperref}
\hypersetup{hidelinks=true}
\usepackage{algorithm,algorithmic}
\usepackage[T1]{fontenc}
\usepackage{hyperref, times, cite}
\usepackage{url}
\usepackage{mathtools}
\usepackage{bm, bbm}
\usepackage{cool}
\usepackage{amssymb,amsfonts,amsmath,amsthm,amscd,dsfont,mathrsfs}
\usepackage{graphicx,float,psfrag,epsfig,amssymb}
\usepackage{wrapfig}
\usepackage{relsize}
\usepackage{color}
\usepackage{pict2e}
\usepackage{multicol}
\usepackage{algorithm}
\usepackage{algorithmic}
\usepackage{caption}
\usepackage{nameref}
\usepackage{enumerate}
\usepackage{stackrel}
\usepackage{caption}
\usepackage{subcaption}

\usepackage{amsthm}

\begin{document}

\title{On Conditional Independence Graph Learning From Multi-Attribute Gaussian Dependent Time Series}
\author{ Jitendra K.\ Tugnait 
\thanks{J.K.\ Tugnait is with the Department of 
Electrical \& Computer Engineering,
200 Broun Hall, Auburn University, Auburn, AL 36849, USA. 
Email: tugnajk@auburn.edu . }
\thanks{This work was supported by the National Science Foundation under Grant CCF-2308473.}}

\maketitle

\renewcommand{\algorithmicrequire}{\textbf{Input:}}
\renewcommand{\algorithmicensure}{\textbf{Output:}}

\begin{abstract}
Estimation of the conditional independence graph (CIG) of high-dimensional multivariate Gaussian time series from multi-attribute data is considered. Existing methods for graph estimation for such data are based on single-attribute models where one associates a scalar time series with each node. In multi-attribute graphical models, each node represents a random vector or vector time series. In this paper we provide a unified theoretical analysis of multi-attribute graph learning for dependent time series using a penalized log-likelihood objective function formulated in the frequency domain using the discrete Fourier transform of the time-domain data. We consider both convex (sparse-group lasso) and non-convex (log-sum and SCAD group penalties) penalty/regularization functions. We establish  sufficient conditions in a high-dimensional setting for consistency (convergence of the inverse power spectral density to true value in the Frobenius norm), local convexity when using non-convex penalties, and graph recovery. We do not impose any incoherence or irrepresentability condition for our convergence results.  We also empirically investigate selection of the tuning parameters based on the Bayesian information criterion, and illustrate our approach using numerical examples utilizing both synthetic and real data.
\end{abstract}


\maketitle

\section{INTRODUCTION} \label{intro}
\IEEEPARstart{G}{raphical}  models are a useful tool for analyzing multivariate data where conditional independence plays an important role \cite{Whittaker1990, Lauritzen1996, Buhlmann2011, Meinshausen2006}. Let ${\cal G} = \left( V, {\cal E} \right)$ denote a graph with a set of $p$ vertices (nodes) $V = \{1,2, \cdots , p\} =[p]$, and a corresponding set of (undirected) edges ${\cal E} \subseteq [p] \times [p]$. Consider a stationary, zero-mean,  $p-$dimensional multivariate Gaussian time series ${\bm x}(t)$, $t=0, \pm 1, \pm 2, \cdots $, with $i$th component $x_i(t)$, and correlation (covariance) matrix function ${\bm R}_{xx}( \tau ) = \mathbb{E} \{ {\bm x}(t + \tau) {\bm x}^T(t ) \}$, $ \tau = 0, \pm 1,  \cdots  $. Given $\{ {\bm x}(t) \}$, in the corresponding graph ${\cal G}$, each component series $\{ x_i(t) \}$ is represented by a node ($i$ in $V$), and associations between components $\{ x_i (t) \}$ and $\{ x_j(t) \}$ are represented by edges between nodes $i$ and $j$ of ${\cal G}$. In a conditional independence graph (CIG), there is no edge between nodes $i$ and $j$ (i.e., $\{ i,j \} \not\in {\cal E}$) if and only if (iff) $x_i(t)$ and $x_j(t)$ are conditionally independent given the remaining $p$-$2$ scalar series $x_\ell(t)$, $\ell \in [p]$, $\ell \neq i$, $\ell \neq j$. (This is a generalization of the CIG for random vectors where $\{ i,j \} \not\in {\cal E}$ iff $[{\bm \Omega}]_{ij} = 0$ (${\bm \Omega} = (E\{ {\bm x}(t) {\bm x}^\top(t) \})^{-1}$) \cite{Meinshausen2006, Banerjee2008, Dahlhaus2000}.)  

Denote the power spectral density (PSD) matrix of $\{ {\bm x}(t) \}$ by ${\bm S}_x(f)$, where ${\bm S}_x(f) = \sum_{\tau = -\infty}^{\infty}  {\bm R}_{xx}( \tau ) e^{-\iota 2 \pi f \tau}$ and $\iota = \sqrt{-1}$.  In \cite{Dahlhaus2000} it was shown that conditional independence of two time series components given all other components of the time series, is encoded by zeros in the inverse PSD, that is, $\{ i,j \} \not\in {\cal E}$ iff the $(i,j)$-th element of ${\bm S}_x^{-1}(f)$,  $[{\bm S}_x^{-1}(f)]_{ij} = 0$ for every $f$. Hence one can use estimated inverse PSD of observed time series to infer the associated graph. In \cite{Dahlhaus2000} the low-dimensional case is addressed.  Nonparametric frequency-domain approaches for graph estimation in high-dimensional settings (sample size $n$ is less than or of the order of $p$) have been considered in \cite{Jung2015a} using a group-lasso penalty,  and in \cite{Tugnait18c, Tugnait20,  Tugnait22c} using a sparse-group lasso penalty. The focus of this paper is on high-dimensional settings where the number of graph nodes $p$ (e.g., time series dimension) is smaller than or comparable to the data sample size $n$ \cite{Wainwright2019}. In particular, in a high-dimensional setting, as $n \rightarrow \infty$, $\frac{p}{n} \rightarrow c > 0$ for some constant $c$, instead of $\frac{p}{n} \rightarrow  0$ as in classical low-dimensional statistical analysis framework \cite[Chapter 1]{Wainwright2019}. Such models for the i.i.d.\ $\{\bm x(t) \}$ case have been extensively studied \cite{Meinshausen2006, Banerjee2008, Wainwright2019}. If $\frac{p}{n} \ll 1$, we use the term low-dimensional for such cases in this paper.  A sparse-group non-convex log-sum penalty is investigated in \cite{Tugnait22d} to regularize the problem considered in \cite{Tugnait22c}, motivated by \cite{Candes2008}. Refs.\ \cite{Jung2015a, Tugnait20, Tugnait22c} provide performance analysis and guarantees. 

Parametric modeling (autoregressive (AR) or autoregressive moving average (ARMA) models) based approaches in low-dimensional settings for CIG estimation for time series  are discussed in \cite{Avventi2013, Alpago2018, Songsiri2010, You2022}, among others. These papers are focused on algorithm development and they do not provide any performance guarantees (such as \cite[Theorem 1]{Tugnait22c} or Theorem 1 in this paper). Compared with this paper or \cite{Tugnait22c} where the high-dimensional case is considered, \cite{Avventi2013, Alpago2018, Songsiri2010, You2022} consider a low-dimensional setting. For instance, in the simulation example 1 of \cite{You2022}, one has a 10-dimensional ARMA model implying a 10-node graph ($p=10$ in our notation) while the data sample size used to illustrate the performance of their algorithm is 1024 ($n=1024$ in our notation), leading to $p/n = 0.0098 \ll 1$. In contrast, in the synthetic data example in \cite[Sec.\ 6.1]{Tugnait22c}, one has $p=128$ and $n \in \{128, 256, 512, 1024, 2048\}$, leading to $p/n \in \{1, 0.5, 0.25, 0.125, 0.0625 \}$. Statistical analysis in the high-dimensional case requires a different set of analytical tools \cite{Buhlmann2011, Wainwright2019}. Estimation of ARMA models for high-dimensional Gaussian time series without considering graphical modeling aspects is discussed in \cite{Basu2015} where in \cite[Example 1]{Basu2015}, one has $p=200$ with varying values of sample size $n$ with some values of $n < 50$.

In many applications, there may be more than one random variable (or scalar time series) associated with a node. This class of graphical models has been called multi-attribute graphical models in \cite{Kolar2014, Tugnait21a} where a high-dimensional setting is considered, and vector graphs or networks in \cite{Marjanovic18, Yue20, Sundaram20} where a low-dimensional setting is considered. In a gene regulatory network, one may have different molecular profiles available for a single gene, such as protein, DNA and RNA. Since these molecular profiles are on the same set of biological samples, they constitute multi-attribute data for gene regulatory graphical models in \cite{Kolar2014}. The motivation for vector graphical models considered in \cite{Marjanovic18, Yue20, Sundaram20} is network analysis for human fMRI data. In this paper in Sec.\ \ref{NE}-\ref{NEreal}, we model air-quality and meteorological data acquired at different monitoring stations in Beijing \cite{Zhang2017, Chen2015} as multi-attribute data, with measurements of each variable at $m$ stations modeled as $m$ attributes. Such graphical models have been considered in the literature only for random vectors (i.e., observations originate from an i.i.d.\ random sequence), not for time series graphical models. The objective of this paper is to fill this gap. Additionally, we consider both convex (sparse-group lasso \cite{Friedman2010a, Simon2013}) and non-convex (log-sum \cite{Candes2008} and Smoothly Clipped Absolute Deviation (SCAD) \cite{Fan2001, Lam2009}) penalty functions. It is well-known that use of non-convex penalties can yield more accurate results compared to the lasso penalty, i.e., they can produce sparse set of solution like lasso, and approximately unbiased coefficients for large coefficients, unlike lasso \cite{Fan2001, Candes2008, Lam2009}. This motivates consideration of the SCAD and log-sum penalties (in addition to the lasso penalty) in this paper. As noted earlier,  a sparse-group non-convex log-sum penalty is investigated in \cite{Tugnait22d} to regularize the single-attribute problem considered in \cite{Tugnait22c} where it is shown empirically that the log-sum penalty significantly outperforms the lasso penalty. Hence the interest in non-convex penalties in this paper.

\subsection{RELATED WORK}
There appears to be no prior reported work on graphical modeling for multi-attribute dependent time series in high-dimensional settings. Prior work on graphical modeling for single-attribute dependent time series in low-dimensional settings is concerned with testing whether $\{ i,j \} \in {\cal E}$ for all possible edges in the graph, based on some nonparametric frequency-domain test statistic such as partial coherence \cite{Dahlhaus2000, Matsuda2006, Wolstenholme2015, Luftman2016, Tugnait19d} which requires estimates of ${\bm S}_x(f)$. These approaches do not scale to high dimensions where $p$ is comparable to or larger than the sample size $n$. As an alternative to nonparametric modeling of time series, parametric graphical models utilizing  (Gaussian) vector AR (VAR) process models of ${\bm x}(t)$ have been proposed in \cite{Eichler2006, Eichler2012, Songsiri2009, Songsiri2010} and ARMA process (and related) models may be found in \cite{Avventi2013, Alpago2018, You2022}, but these approaches are suitable only for low-dimensional settings as discussed earlier. These approaches do not address the multi-attribute case. Graphical modeling for single-attribute dependent time series in high-dimensional settings has been considered using nonparametric frequency-domain approaches in \cite{Jung2015a, Tugnait18c, Tugnait20,  Tugnait22c} with convex lasso-related regularization and in \cite{Tugnait22d} with non-convex log-sum regularization. A time-domain approach with log-sum penalty may be found in \cite{Tugnait2022s}

Multi-attribute graphical modeling in high-dimensional setting given i.i.d.\ data has been addressed in \cite{Kolar2014, Tugnait21a} using convex lasso-related regularization and in \cite{Tugnait2021d} using non-convex SCAD penalty. When convex regularization is used, the overall optimization problem is convex where a global optimum solution is guaranteed, whereas with non-convex penalties, one can obtain only a local optimum.  

This paper builds on the work reported in \cite{Tugnait22c}. A detailed comparison between this paper and \cite{Tugnait22c} (also \cite{Tugnait22d}) is given later in Remark 4 in Sec.\ \ref{consist} after we have introduced all the technical details facilitating the comparison.

\subsection{OUR CONTRIBUTIONS} 
In this paper we provide a unified theoretical analysis of multi-attribute graph learning for dependent time series using a penalized log-likelihood objective function in the frequency domain. We consider the convex sparse-group lasso as well as the non-convex log-sum and SCAD group penalties. The non-convex optimization problem (when using non-convex penalties) is solved via iterative convex optimization, based on a local-linear approximation (LLA) \cite{Zou2008, Lam2009} to the non-convex penalty  and an alternating direction method of multipliers (ADMM) method. The ADMM method used in this paper follows \cite{Tugnait22c} and differences between \cite{Tugnait22c} and this paper are explained later in Sec.\ \ref{opt} and in Remark 4 in Sec.\ \ref{consist}. We establish  sufficient conditions in a high-dimensional setting for consistency (convergence of the inverse power spectral density to true value in the Frobenius norm) in Theorem 1, local convexity when using non-convex penalties in Theorem 2, and graph recovery in Theorem 3. We do not impose any incoherence or irrepresentability condition for our Theorems 1-3 (see Remark 3 in Sec.\ \ref{consist}).  We illustrate our approach using numerical examples utilizing both synthetic and real (Beijing air-quality \cite{Zhang2017, Chen2015}) data.

A preliminary version of this paper appears in a conference paper \cite{Tugnait2024} where proofs of Theorems 1 and 3 and Lemma 1 are not given, and only a sketch of proof of Theorem 2 appears. Theorem 1 in \cite{Tugnait2024} has an error. Synthetic data examples are different in this paper and \cite{Tugnait2024}.

\subsection{OUTLINE AND NOTATION} \label{outnot}
The rest of the paper is organized as follows.  The underlying system model and the resulting log-likelihood formulation of the problem are presented in Sec.\ \ref{SM}.  The convex and non-convex penalty functions and their properties (based on \cite{Loh2017}), and the resulting penalized negative log-likelihood function is discussed in Sec.\ \ref{PLL}. A solution to the non-convex optimization problem is provided in Sec.\ \ref{opt}. Selection of the tuning parameters based on BIC is presented in Sec.\ \ref{opt}-\ref{bic}. In Sec.\ \ref{consist} we provide a theoretical analysis of the proposed approach, resulting in Theorems 1-3. Numerical results are presented in Sec.\ \ref{NE} and proofs of Theorems 1, 2 and 3 are given in the two appendices.

 The superscripts $\ast$, $\top$ and $H$ denote the complex conjugate, transpose and  Hermitian (conjugate transpose) operations, respectively, and the sets of real, positive real and complex numbers are denoted by $\mathbb{R}$, $\mathbb{R}_+$ and $\mathbb{C}$, respectively. Given  ${\bm A} \in \mathbb{C}^{p \times p}$, we use $\phi_{\min }({\bm A})$, $\phi_{\max }({\bm A})$, $|{\bf A}|$, $\mbox{tr}({\bm A})$ and $\mbox{etr}({\bm A})$ to denote the minimum eigenvalue, maximum eigenvalue, determinant, trace, and exponential of trace of ${\bm A}$, respectively. We use ${\bm A} \succeq 0$ and ${\bm A} \succ 0$ to denote that Hermitian ${\bm A}$ is positive semi-definite and positive definite, respectively, and ${\bm I}_p$ is the $p \times p$ identity matrix. For ${\bm B} \in \mathbb{C}^{p \times q}$, we define  the operator norm, the Frobenius norm and the vectorized $\ell_1$ norm, respectively, as $\|{\bm B}\| = \sqrt{\phi_{\max }({\bm B}^H  {\bm B})}$, $\|{\bm B}\|_F = \sqrt{\mbox{tr}({\bm B}^H  {\bm B})}$ and $\|{\bm B}\|_1 = \sum_{i,j} |B_{ij}|$, where $B_{ij}$ is the $(i,j)$-th element of ${\bm B}$, also denoted by $[{\bm B}]_{ij}$.  For vector ${\bm \theta} \in \mathbb{C}^p$, we define $\| {\bm \theta} \|_1 = \sum_{i=1}^p |\theta_i|$ and $\| {\bm \theta} \|_2 = \sqrt{\sum_{i=1}^p |\theta_i|^2}$, and we also use $\| {\bm \theta} \|$ for $\| {\bm \theta} \|_2$. The Kronecker product of matrices ${\bm A}$ and ${\bm B}$ is denotes by ${\bm A} \otimes {\bm B}$. Given ${\bm A} \in \mathbb{C}^{p \times p}$, ${\bm A}^+ = \mbox{diag}({\bm A})$ is a diagonal matrix with the same diagonal as ${\bm A}$, and  ${\bm A}^- = {\bm A} - {\bm A}^+$ is ${\bm A}$ with all its diagonal elements set to zero. Given ${\bm A} \in \mathbb{C}^{n \times p}$, column vector $\mbox{vec}({\bm A}) \in \mathbb{C}^{np}$ denotes the vectorization of ${\bm A}$ which stacks the columns of the matrix ${\bm A}$. The notation ${\bm x} \sim {\mathcal N}_c( {\bf m}, {\bm \Sigma})$ denotes a complex random vector  ${\bm x}$ that is circularly symmetric (proper), complex Gaussian with mean ${\bm m}$ and covariance ${\bm \Sigma}$, and ${\bm x} \sim {\mathcal N}_r( {\bf m}, {\bm \Sigma})$ denotes real-valued Gaussian ${\bm x}$ with mean ${\bm m}$ and covariance ${\bm \Sigma}$.

\section{SYSTEM MODEL} \label{SM} Consider $p$ jointly Gaussian, zero-mean stationary, vector sequences $\{ {\bm z}_i (t) \}_{t \in \mathbb{Z}}$, ${\bm z}_i(t) \in \mathbb{R}^m$, $i \in [p]$.  In a multi-attribute time series graphical model, we associate $\{ {\bm z}_i (t) \}_{t \in \mathbb{Z}}$ with the $i$th node of an undirected graph ${\cal G} = \left( V, {\cal E} \right)$ where $V =[p]$ is the set of $p$ nodes (vertices) and ${\cal E} \subseteq V \times V $ is the set of undirected edges that describe the conditional dependencies among the $p$ sequences $\{ \{ {\bm z}_i (t) \}_{t \in \mathbb{Z}}, \; i \in V \}$. Similar to the scalar case ($m=1$), edge $\{i,j\} \not\in {\cal E}$ iff the sequences $\{ {\bm z}_i (t) \}$ and $\{ {\bm z}_j (t) \}$ are conditionally independent given the remaining $p-2$ vector sequences $\{ {\bm z}_\ell (t) \}$, $\ell \in V \backslash \{i,j\}$.

Define the $mp$-dimensional sequence
\begin{equation} \label{neweq10}
   {\bm x}(t) = \big[ {\bm z}_1^\top(t) , \;  {\bm z}_2^\top(t) , \; \cdots  , \; {\bm z}_m^\top(t) \big]^\top 
	  \in \mathbb{R}^{mp} \, .
\end{equation}
Associate $\{ {\bm x} (t) \}_{t \in \mathbb{Z}}$ with an enlarged graph $\bar{\cal G} = \left( \bar{V}, \bar{\cal E} \right)$ where $\bar{V} =[mp]$ and $\bar{\cal E} \subseteq \bar{V} \times \bar{V}$. The $\ell$th component of $\{ {\bm z}_j (t) \}$, denoted by $\{ [{\bm z}_j]_\ell (t) \}$, associated with the node $j$ of ${\cal G}$, is the scalar sequence $\{ x_q(t) \}$, $x_q = [{\bm x}]_q$, $q=(j-1)m+\ell$, $j \in [p]$ and $\ell \in [m]$. The scalar sequence $\{ x_q(t) \}$ is associated with node $q$ of enlarged graph $\bar{\cal G}$. Corresponding to the edge $\{j,k\} \in V \times V$ in ${\cal G}$, there are $m^2$ edges $\{q,r\} \in \bar{V} \times \bar{V}$ in $\bar{\cal G}$ where $q=(j-1)m+u$ and $r=(k-1)m+v$ with $u,v \in [m]$. 

As in Sec.\ \ref{intro}, denote the power spectral density (PSD) matrix of $\{ {\bm x}(t) \}$ by ${\bm S}_x(f)$.  Here $f$ is the normalized frequency, in Hz. Given a matrix ${\bm A} \in \mathbb{C}^{(mp) \times (mp)}$, we use ${\bm A}^{(jk)}$ to denote the $m \times m$ submatrix of ${\bm A}$ whose $(u,v)$th element is given by
\begin{equation} \label{neweq12}
   [{\bm A}^{(jk)}]_{uv} = [{\bm A}]_{(j-1)m+u, (k-1)m+v} \, , \quad u,v \in [m].
\end{equation}
By \cite[Theorem 2.4]{Dahlhaus2000}, in the CIG ${\cal G} = \left( V, {\cal E} \right)$ of the multi-attribute time series $\{ {\bm x} (t) \}_{t \in \mathbb{Z}}$ originating via (\ref{neweq10}), we have
\begin{equation} \label{neweq14}
   \{j,k\} \not\in {\cal E} \;\; \Leftrightarrow \;\; \left({\bm S}_x^{-1}(f)\right)^{(jk)} \equiv {\bm 0}
\end{equation}
provided ${\bm S}_x(f) \succ {\bm 0}$ $\forall f$. (Note that while most of the discussion and all of the numerical results in \cite{Dahlhaus2000} pertain to scalar time series per node, the theory is shown to apply to vector series per node also.)

\subsection{PROBLEM FORMULATION} \label{BPF1}
We observe a finite-duration segment $\{ {\bm x}(t) \}_{t=0}^{n-1}$ of a realization of an $mp-$dimensional stationary Gaussian sequence $\{ {\bm x}(t) \}_{t \in \mathbb{Z}}$. Our objective is to first estimate the inverse PSD ${\bm S}_x^{-1}(f)$ at distinct frequencies, and then select the edge $\{j,k\}$ in the graphical model ${\cal G}$ based on whether or not $\left({\bm S}_x^{-1}(f)\right)^{(jk)} = {\bm 0}$ for every $f$. The single attribute case ($m=1$) has been discussed in \cite{Tugnait22c} with sparse-group lasso penalty and in \cite{Tugnait22d} with sparse-group log-sum penalty. Since for a real-valued time series, ${\bm S}_x(f) = {\bm S}_x^H(-f)$, and ${\bm S}_x(f)$ is periodic in $f$ with period one, knowledge of ${\bm S}_x(f)$ in the interval $[0,0.5]$ completely specifies ${\bm S}_x(f)$ for other values of $f$. Hence, it is enough to check if $\left({\bm S}_x^{-1}(f)\right)^{(jk)} = {\bm 0}$ for every $f \in [0,0.5]$.

Given $\{ {\bm x}(t) \}_{t=0}^{n-1}$, define the (normalized) DFT ${\bm d}_x(f_\ell)$ of ${\bm x}(t)$, ($\iota = \sqrt{-1}$),
\begin{equation} 
   {\bm d}_x(f_\ell) =  \frac{1}{\sqrt{n}} \sum_{t=0}^{n-1} {\bm x}(t) \exp \left( - \iota 2 \pi f_\ell t \right) \, ,
	   \label{app1eq50} \\
\end{equation}
where
\begin{equation} \label{app1eq50a}
	   f_\ell =  \ell/n, \quad \ell=0,1, \cdots , n-1. 
\end{equation}
Since $\{ {\bm x}(t) \}$ is Gaussian, so is ${\bm d}_x(f_\ell)$. As discussed in \cite{Tugnait22c}, the set of complex-valued random vectors $\{{\bm d}_x(f_\ell)\}_{\ell=0}^{n/2}$, $n$ even, is a sufficient statistic for any statistical inference problem, including our problem of estimation of inverse PSD.

We need the following assumption in order to invoke \cite[Theorem 4.4.1]{Brillinger}, used extensively later.
\begin{itemize}
\setlength{\itemindent}{0.1in}
\item[(A1)] The $mp-$dimensional time series $\{ {\bm x}(t) \}_{t \in \mathbb{Z}}$ is zero-mean stationary and Gaussian, satisfying 
\[
    \sum_{\tau = -\infty}^\infty | [{\bm R}_{xx}( \tau )]_{k \ell} | < \infty \mbox{ for every } 
		  k, \ell \in \bar{V}  \, .
\] 
\end{itemize}
It follows from \cite[Theorem 4.4.1]{Brillinger} that under assumption (A1), asymptotically (as $n \rightarrow \infty$), ${\bm d}_x(f_\ell)$, $\ell \in [(n/2)-1]$, ($n$ even), are independent proper (i.e., circularly symmetric), complex Gaussian ${\mathcal N}_c( {\bf 0}, {\bm S}_x(f_\ell))$ random vectors, respectively. Also, asymptotically, ${\bm d}_x(f_0)$ and ${\bm d}_x(f_{n/2})$, ($n$ even), are independent real Gaussian ${\mathcal N}_r( {\bf 0}, {\bm S}_x(f_0))$ and ${\mathcal N}_r( {\bf 0}, {\bm S}_x(f_{n/2}))$ random vectors, respectively, independent of ${\bm d}_x(f_\ell)$, $\ell \in [(n/2)-1]$. We will ignore these two frequency points $f_0$ and $f_{n/2}$.

Define 
\begin{equation} \label{eqth1_150a}
    {\bm D} = \left[ {\bm d}_x(f_1) \;  \cdots \; {\bm d}_x(f_{(n/2)-1}) \right]
	   \in \mathbb{C}^{(mp) \times ((n/2)-1) } \, .
\end{equation}
We assume that ${\bm S}_x(f_\ell)$ is locally smooth (a standard assumption in PSD estimation \cite{Brillinger}), so that ${\bm S}_x(f_\ell)$ is (approximately) constant over $K=2m_t+1$ consecutive frequency points. Pick
\begin{equation} \label{window}
  \tilde{f}_k = \frac{(k-1)K+m_t+1}{n}, \;\; \quad k=1,2, \cdots , M \, ,
\end{equation}
\begin{equation} \label{windowM}
	 M = \Big\lfloor \big( \frac{n}{2}-m_t-1 \big) / K  \Big\rfloor \, ,
\end{equation}
leading to $M$ equally spaced frequencies $\tilde{f}_k$ in the interval $(0,0.5)$, at intervals of $K/n$. We state the local smoothness assumption as assumption (A2).
\begin{itemize}
\setlength{\itemindent}{0.1in}
\item[(A2)] Assume that for $\ell = -m_t, -m_t+1, \cdots , m_t$,
\begin{align}  
   {\bm S}_x & (\tilde{f}_{k,\ell}) =  {\bm S}_x(\tilde{f}_k) \,  , \label{eqth1_160}  \\
	 \mbox{ where } \;  &
	  \tilde{f}_{k,\ell} = \big((k-1)K+m_t+1 + \ell \big)/n \, .
\end{align}
\end{itemize}

Under assumptions (A1)-(A2), the joint pdf of ${\bm D}$ is 
\begin{align}  
   f_{{\bm D}}({\bm D})  = & \prod_{k=1}^{M}  \left[ \prod_{\ell = -m_t}^{m_t}
	   \frac{\exp \left(- g_{kl} - g_{kl}^\ast  \right) }
		 { \pi^{mp} \, | {\bm S}_x^{-1}(\tilde{f}_k) |^{1/2} \, | {\bm S}_x^{-\ast}(\tilde{f}_k) |^{1/2} }
		     \right] \, ,  \label{eqth1_162} \\
  g_{kl} = & \frac{1}{2}{\bm d}_x^H(\tilde{f}_{k,\ell}) 
          {\bm S}_x^{-1}(\tilde{f}_k)  {\bm d}_x(\tilde{f}_{k,\ell})  \, , \label{eqth1_162b} 
\end{align}
where ${\bm A}^{-\ast}$ stands for $({\bm A}^{-1})^\ast$. Parametrizing in terms of the inverse PSD matrix $\bm{\Phi}_k := {\bm S}_x^{-1}(\tilde{f}_k)$, the negative log-likelihood, up to some irrelevant constants, is given by
\begin{align} 
   & - \ln  f_{{\bm D}}({\bm D})  \propto {\cal L}({\bm \Omega}) \\
					&	:= \sum_{k=1}^M \frac{1}{2} \left[ -\ln (|\bm{\Phi}_k|) -\ln( |\bm{\Phi}_k^\ast|) 
		        + {\rm tr} \left( \hat{\bm S}_k \bm{\Phi}_k  +
						     \hat{\bm S}_k^\ast \bm{\Phi}_k^\ast \right)  \right]   	 \label{eqth2_10}
\end{align}
where
\begin{equation}  \label{neqn100}
  {\bm \Omega} =[ {\bm \Phi}_1 \, , \;  {\bm \Phi}_2  \, , \; \cdots \;  \, , {\bm \Phi}_M ] \in \mathbb{C}^{(mp) \times (mpM)} \, ,
\end{equation}
\begin{equation}  \label{moresm} 
	\hat{\bm S}_k =  \frac{1}{K} \sum_{\ell= - m_t}^{m_t} {\textbf d}_x(\tilde{f}_{k,\ell}) 
		 {\textbf d}_x^H(\tilde{f}_{k,\ell}) \, . 
\end{equation}
Note that $\hat{\bm S}_k$ represents PSD estimator at frequency $\tilde{f}_{k}$ using unweighted frequency-domain smoothing \cite{Brillinger}.

Our objective is to estimate ${\bm \Omega}$ given $\{ {\bm x}(t) \}_{t=0}^{n-1}$, and to infer the underlying CIG based on estimated ${\bm \Omega}$.

\section{PENALIZED NEGATIVE LOG-LIKELIHOOD} \label{PLL}
To enforce sparsity and to make the problem well-conditioned (when $K < p$), as in \cite{Tugnait22c}, we propose to minimize a penalized version  $\bar{\cal L}({\bm \Omega})$ of ${\cal L}({\bm \Omega})$ where we penalize (regularize) at both element-wise and group-wise.  We have
\begin{align}  
   \bar{\cal L}({\bm \Omega}) & =  {\cal L}({\bm \Omega}) + \alpha P_e({\bm \Omega}) 
	   + (1-\alpha) P_g({\bm \Omega}) , \label{eqth2_20} \\
	   P_e({\bm \Omega}) &  =  \sum_{k=1}^M \; \sum_{i \ne j}^{mp} 
		  \rho_\lambda \left(  [ {\bm{\Phi}}_k ]_{ij}  \right)  , \label{eqth2_20a} \\
		P_g({\bm \Omega}) &  = m \sqrt{M} \, \sum_{ q \ne \ell}^p \; 
		    \rho_\lambda \left( \| {\bm \Omega}^{(q \ell M)} \|_F \right) \label{eqth2_20b} 
\end{align}
where ${\bm \Omega}^{(q \ell M)} \in \mathbb{C}^{m \times (mM)}$ is defined as
\begin{align}  
{\bm \Omega}^{(q \ell M)} & := [ {\bm{\Phi}}_1^{(q \ell)} , \; {\bm{\Phi}}_2^{(q \ell)} ,
   \; \cdots ,\; {\bm{\Phi}}_M^{(q \ell)}]
       \, , \label{eqth2_20c}
\end{align}
${\bm{\Phi}}_i^{(q \ell)}$, $i \in [M]$, is defined as in (\ref{neweq12}), $\lambda > 0$, $\alpha \in [0,1]$, $m \sqrt{M}$ in (\ref{eqth2_20b}) reflects the number of group variables \cite{Yuan2007}, and for $u \in \mathbb{R}$, $\rho_\lambda(u)$ is a penalty function that is function of $|u|$. In (\ref{eqth2_20a}), the penalty term is applied to each off-diagonal element of ${\bm{\Phi}}_k$ and in (\ref{eqth2_20b}), the penalty term is applied to the off-block-diagonal group of $m^2M$ terms via ${\bm \Omega}^{(q \ell M)}$, defined in (\ref{eqth2_20c}). The parameter $\alpha \in [0,1]$ ``balances'' element-wise and group-wise penalties \cite{Friedman2010a, Tugnait22c}

The following penalty functions are considered:
\begin{itemize}
\item {\it Lasso}. For some $\lambda > 0$,  
\begin{equation}  \label{lasso} 
        \rho_\lambda(u) = \lambda |u| , \quad u \in \mathbb{R} \, .
\end{equation}
\item {\it Log-sum}. For some $\lambda > 0$ and $1 \gg \epsilon > 0$, 
\begin{equation}  \label{logsum}
        \rho_\lambda(u) = \lambda \epsilon \, \ln \left( 1 + \frac{|u|}{\epsilon} \right) \, .
\end{equation}
\item {\it Smoothly Clipped Absolute Deviation (SCAD)}. For some $\lambda > 0$ and $a > 2$, 
\begin{equation}  \label{scad}
   \rho_\lambda(u) = \left\{ \begin{array}{ll}
	   \lambda | u | & \mbox{for } |u| \le \lambda \\
		\frac{2 a \lambda | u |- | u |^2 - \lambda^2}{2 (a-1)}
		      & \mbox{for } \lambda < |u| < a \lambda \\
		\frac{\lambda^2 (a+1)}{2} &  \mbox{for } |u| \ge a \lambda \, . \end{array} \right. 
		 % \label{eqscad} 
\end{equation}
\end{itemize}
In the terminology of \cite{Loh2017}, all of the above three penalties are ``$\mu$-amenable'' for some $\mu \ge 0$. As defined in \cite[Sec.\ 2.2]{Loh2017}, $\rho_\lambda(u)$ is $\mu$-amenable for some $\mu \ge 0$ if 
\begin{itemize}
\item[(i)] The function $\rho_\lambda(u)$ is symmetric around zero, i.e., $\rho_\lambda(u) = \rho_\lambda(-u)$ and $\rho_\lambda(0) = 0$.
\item[(ii)] The function $\rho_\lambda(u)$ is non-decreasing on $\mathbb{R}_+$.
\item[(iii)] The function $\rho_\lambda(u)/u$ is non-increasing on $\mathbb{R}_+$.
\item[(iv)] The function $\rho_\lambda(u)$ is differentiable for $u \ne 0$.
\item[(v)] The function $\rho_\lambda(u) +\frac{\mu}{2} u^2$ is convex, for some $\mu \ge 0$.
\item[(vi)] $\lim_{u \rightarrow 0^+} \rho^\prime(u) = \lambda$ where $\rho^\prime(u) := \frac{d \rho_\lambda (u)}{du}$.
\end{itemize}
It is shown in \cite[Appendix A.1]{Loh2017}, that all of the above three penalties are $\mu$-amenable with $\mu = 0$ for Lasso and $\mu =1/(a-1)$ for SCAD. In \cite{Loh2017} the log-sum penalty is defined as $\rho_\lambda(u) = \ln (1+\lambda |u|)$ whereas in \cite{Candes2008}, it is defined as $\rho_\lambda(u) = \lambda \, \ln \left( 1 + \frac{|u|}{\epsilon} \right)$. We follow  \cite{Candes2008} but modify it so that property (vi) in the definition of $\mu$-amenable penalties holds. In our case $\mu = \frac{\lambda}{\epsilon}$ for the log-sum penalty since $\frac{d^2 \rho_\lambda (u)}{du^2} = - \lambda \epsilon /(\epsilon + |u|)^2$ for $u \ne 0$.

The above three penalty functions also have the following properties:
\begin{itemize}
\item[(vii)] For some $C_\lambda > 0$ and $\delta_\lambda > 0$, the function $\rho_\lambda(u)$ has a lower bound 
\begin{equation}  \label{prop7} 
        \rho_\lambda(u) \ge C_\lambda |u| \mbox{  for  } |u| \le \delta_\lambda \, .
\end{equation}
\item[(viii)] $\frac{d \rho_\lambda (u)}{d |u|} \le \lambda$ for $u \ne 0$.
\end{itemize}
Property (viii) is straightforward to verify. For Lasso,  $C_\lambda = \lambda$ and $\delta_\lambda = \infty$. For SCAD, $C_\lambda = \lambda$ and $\delta_\lambda = \lambda$. Since $\ln (1+x) \ge x/(1+x)$ for $x > -1$, we have $\ln (1+x) \ge x/C_1$ for $0 \le x \le C_1-1$, $C_1 > 1$. Take $C_1 =2$. Then log-sum $\rho_\lambda(u) \ge \frac{\lambda}{2} |u|$ for any $|u| \le \epsilon$, leading to $C_\lambda = \frac{\lambda}{2}$ and $\delta_\lambda =\epsilon$. We may and will take $C_\lambda = \frac{\lambda}{2}$ for lasso and SCAD penalties as well.

\section{OPTIMIZATION}  \label{opt}
For non-convex $\rho_\lambda(u)$, we will use a local linear approximation (LLA) as in \cite{Zou2008, Lam2009}, to yield 
\begin{equation}
  \rho_{\lambda}(u) \approx \rho_{\lambda}(|u_0|) 
	 + \rho_\lambda^\prime(|u_0|) (|u| - |u_0|)
	 \, \Rightarrow \, \rho_\lambda^\prime(|u_0|) |u| \, ,
\end{equation}
where $u_0$ is an initial guess, $\rho_\lambda^\prime(|u_0|) = \lambda \epsilon /(|u_0| +\epsilon)$ for LSP, and for SCAD, $\rho_\lambda^\prime(|u_0|) = \lambda$ for $|u| \le \lambda$, $=\frac{ a \lambda - | u |}{ a-1}$ for $\lambda < |u| < a \lambda$, and $=0$ for $|u| \ge a \lambda$. Therefore, with $u_0$ fixed, we consider only the last term above for optimization w.r.t.\ $u$. By \cite[Theorem 1]{Zou2008}, the LLA provides a majorization of the non-convex penalty, thereby yielding a majorization-minimization approach. In fact, by \cite[Theorem 2]{Zou2008}, the LLA is the best convex majorization of the LSP and SCAD penalties. Thus in LSP, with some initial guess $\bar{\bm{\Phi}}_k$, we replace $\rho_\lambda ( | [ {\bm{\Phi}}_k ]_{ij} | ) \rightarrow \lambda \epsilon /(| [ \bar{\bm{\Phi}}_k ]_{ij} | +\epsilon) =: \lambda_{kij}$ and $\rho_\lambda ( \| {\bm \Phi}^{(q \ell M)} \|_F ) \rightarrow \lambda \epsilon /(\| \bar{\bm \Phi}^{(q \ell M)} \|_F  +\epsilon) =: \lambda_{q \ell M}$, leading an adaptive sparse-group lasso convex problem. The initial guess follows from the solution to lasso-penalized objective function. For SCAD, we have $\lambda_{kij} = \lambda$ for $| [ {\bm{\Phi}}_k ]_{ij} | \le \lambda$, $=(a \lambda - | [ {\bm{\Phi}}_k ]_{ij} |)/(a-1)$ for $ \lambda < | [ {\bm{\Phi}}_k ]_{ij} | \le a \lambda$, and $=0$ otherwise, and similarly for $\lambda_{q \ell M}$. 

With LLA, the objective function is transformed to
\begin{align}  
   \tilde{\cal L}({\bm \Omega}) & =  {\cal L}({\bm \Omega}) + \alpha \tilde{P}_e({\bm \Omega}) 
	   + (1-\alpha) \tilde{P}_g({\bm \Omega}) , \label{admm} \\
	   \tilde{P}_e({\bm \Omega}) &  =  \sum_{k=1}^M \; \sum_{i \ne j}^{mp} 
		   \lambda_{kij} \Big| [ {\bm{\Phi}}_k ]_{ij} \Big|  , \label{admm1} \\
		\tilde{P}_g({\bm \Omega}) &  = m \sqrt{M} \, \sum_{ q \ne \ell}^p \; 
		    \lambda_{q \ell M} \| {\bm \Phi}^{(q \ell M)} \|_F  \, . \label{admm2} 
\end{align}
For lasso, we have $\lambda_{kij} = \lambda$ $\forall k,i,j$ and $\lambda_{q \ell M} = \lambda$ $\forall q, \ell$. 
We follow an ADMM approach, as outlined in \cite{Tugnait22c}, for both lasso and LLA to LSP/SCAD. Consider the scaled augmented Lagrangian \cite{Boyd2004} for this problem after variable splitting, given by 
\begin{align}
 \bar{\cal L}_\rho &(\{ \bm{\Omega} \}, \{{\bm W} \}, \{{\bm U} \} ) =   
   {\cal L}(\{ \bm{\Omega} \})	+ \alpha \tilde{P}_e({\bm W}) 
	    \nonumber \\
		&  + (1-\alpha) \tilde{P}_g({\bm W}) + \frac{\rho}{2} \sum_{k=1}^M \| \bm{\Phi}_k - {\bm W}_k + {\bm U}_k \|^2_F \, , \label{admm0} 
\end{align}
where $\{ {\bm W} \} = \{{\bm W}_k, \; k \in [M] \}$ results from variable splitting where in the penalties we use  ${\bm W}_k$'s instead of ${\bm \Phi}_k$'s, adding the equality constraint ${\bm W}_k={\bm \Phi}_k$, $\{ {\bm U} \} = \{{\bm U}_k, \; k \in [M] \}$ are dual variables,  and $\rho >0$ is the ``penalty parameter'' \cite{Boyd2004}. 

\begin{algorithm} %[H]
\caption{ADMM Algorithm for Solving (\ref{admm})-(\ref{admm0})}
\label{alg0}

\algorithmicrequire{\; PSD estimator $\hat{\bm S}_k$, $k \in [M]$ (computed using (\ref{app1eq50}) and (\ref{moresm})), regularization and penalty parameters $\lambda_{kij}$ ($i,j \in [mp]$, $k \in [M]$), $\lambda_{q \ell M}$ ($q, \ell \in [p]$), $\alpha$ and $\rho=\bar{\rho}$, tolerances $\tau_{abs}$ and $\tau_{rel}$, variable penalty factor $\bar{\mu}$, maximum number of iterations $t_{max}$}. Initial guess $\bar{\bm \Phi}_k$, $k \in [M]$. \\
\algorithmicensure{\;\ Estimated $\hat{\bm \Phi}_k$, $k \in [M]$, and edge-set $\hat{\cal E}$}

\begin{algorithmic}[1] 
\STATE Initialize: ${\bm U}_k^{(0)} = {\bm W}_k^{(0)} = {\bm 0}$, ${\bm \Phi}_k^{(0)} = \bar{\bm \Phi}_k$, $\rho^{(0)} = \bar{\rho}$ 
\STATE converged = \FALSE, $t=0$
\WHILE{converged = \FALSE $\;$ \AND $\;$ $t \le t_{max}$,}
\STATE Let ${\bm V}_k{\bm J}_k{\bm V}_k^H$ denote the eigen-decomposition of Hermitian $\hat{\bm S}_k - \rho^{(t)} \left( {\bm W}_k^{(t)} - {\bm U}_k^{(t)} \right)$, $k \in [M]$, with the diagonal matrix ${\bm J}_k$ consisting of its eigenvalues.  Define a diagonal matrix $\tilde{\bm J}_k$ with $\ell$th diagonal element
$\tilde{\bm J}_{k\ell \ell} = ( -{\bm J}_{k\ell \ell} + \sqrt{ {\bm J}_{k\ell \ell}^2 + 4 \rho^{(t)}  } \, )/(2 \rho^{(t)})$ where ${\bm J}_{k\ell \ell} = [{\bm J}_{k}]_{\ell \ell}$. Set $\bm{\Phi}_k^{(t+1)} = {\bm V}_k \tilde{\bm J}_k {\bm V}_k^H$.
\STATE Define soft thresholding scalar operator $T_{st}(a, \beta) := (1-\beta/|a|)_+ a$ and elementwise matrix soft thresholding operator ${\bm T}_{st}(\bm{A}, \alpha)$, specified by $[{\bm T}_{st}(\bm{A}, \alpha)]_{uv} := T_{st}([\bm{A}]_{uv},\alpha)$, where $(a)_+ := \max(0,a)$ and $u,v \in [m]$. For $k \in [M]$, define ${\bm A}_k = \bm{\Phi}_k^{(t+1)} + {\bm U}_k^{(t)}$ and let $({\bm A}_k)^{(q \ell)} \in \mathbb{C}^{m \times m}$ be defined as in (\ref{neweq12}). Then the diagonal subblocks $({\bm W}_k)^{(q q)} \in \mathbb{C}^{m \times m}$ of ${\bm W}_k$ are updated as ($k \in [M]$)
\begin{align*}  
    [ ({\bm W}_k^{(t+1)})^{(q q)}]_{u v} & = 
	 \left\{ \begin{array}{l}
			    [{\bm A}_k^{(qq)}]_{uu}  \quad \mbox{  if } u=v\\
					\hspace*{-0.1in} T_{st}([{\bm A}_k^{(qq)}]_{uv}, \frac{ \alpha \lambda_{kij}}{\rho^{(t)}}) 
					   \,  \mbox{ if } u \neq v \end{array} \right.
							        %\label{eqth2_4020}
\end{align*}
$q \in [p]$, $\; u,v \in [m]$, $i=(q-1)m+u$, $j=(q-1)m+v$. The off-diagonal $m \times m$ subblocks of ${\bm W}_k$ are updated as 
\begin{align*}  
    ({\bm W}_k^{(t+1)})^{(q \ell)} & = {\bm B} \Big( 1 - \frac{(1-\alpha) m \sqrt{M} \, \lambda_{q \ell M}}
					    {\rho^{(t)} \| {\bm B} \|_F }  \Big)_+ 
							        %\label{eqth2_4020}
\end{align*}
where $m \times m$ ${\bm B}$ has its $(u,v)$th element as $[{\bm B}]_{uv}=T_{st}([{\bm A}_k^{(q \ell)}]_{uv}, \alpha \lambda_{kij}/\rho^{(t)} )$, $i=(q-1)m+u$, $j=(\ell-1)m+v$.
\STATE Dual update ${\bm U}_k^{(t+1)} = {\bm U}_k^{(t)} + \left( {\bm \Phi}_k^{(t+1)} - {\bm W}_k^{(t+1)} \right)$, $k \in [M]$. 
\STATE Check convergence. With $e_1$, $e_2$, $e_3$, ${\bm R}_p^{(t+1)}$, ${\bm R}_d^{(t+1)}$, $\tau_{pri}$ and $\tau_{dual}$ as defined in (\ref{alg100})-(\ref{alg160}), respectively, let $d_p =  \|{\bm R}_p^{(t+1)}\|_F$ and $d_d = \|{\bm R}_d^{(t+1)}\|_F$. 
If $( d_p \le \tau_{pri}) \; \AND \; (d_d \le \tau_{dual})$, set converged = \TRUE .
\STATE Update penalty parameter $\rho$ $\,$ : 
\[
  \rho^{(t+1)} = \left\{ \begin{array}{ll} 2 \rho^{(t)} & \mbox{if  } d_p > \bar{\mu} d_d \\  
	                                         \rho^{(t)} /2 & \mbox{if  } d_d > \bar{\mu} d_p \\
																					 \rho^{(t)} & \mbox{otherwise} \, . \end{array} \right.
\]
We also need to set ${\bm U}^{(t+1)} = {\bm U}^{(t+1)}/2$ for $d_p > \bar{\mu} d_d$ and ${\bm U}^{(t+1)} = 2 {\bm U}^{(t+1)}$ for $d_d > \bar{\mu} d_p$.
\STATE $t \leftarrow t+1$
\ENDWHILE
\STATE Denote the converged inverse PSD estimates as $\hat{\bm \Phi}_k$ and let $\hat{\bm \Omega} = [\hat{\bm \Phi}_1, \; \cdots \;, \hat{\bm \Phi}_M]$. With $\hat{\bm \Omega}^{(q \ell M)}$ as in (\ref{eqth2_20c}), for $q \ne \ell$, if $\|\hat{\bm \Omega}^{(q \ell M)}\|_F > 0$, assign edge $\{ q, \ell\} \in \hat{\cal E}$, else $\{ q, \ell\} \not\in \hat{\cal E}$. 
\end{algorithmic}
\end{algorithm}

The main difference between \cite{Tugnait22c} and this paper is that in \cite{Tugnait22c}, ${\bm W}_k$ and ${\bm \Phi}_k$ are $p \times p$ whereas in this paper, we have ${\bm W}_k$ and ${\bm \Phi}_k$ as $(mp) \times (mp)$ matrices. Therefore, the approach of \cite{Tugnait22c} is applicable after we account for the dimension difference, and additionally, for the fact that $P_g({\bm W})$ and $P_g({\bm \Omega})$ are penalized slightly differently in the two papers (the factor $m \sqrt{M}$ is missing from \cite{Tugnait22c}). See \cite{Tugnait22c} for further details. For non-convex penalties (not considered in \cite{Tugnait22c}), we have an iterative solution: first solve with lasso penalty, then use the LLA formulation and solve the resulting adaptive lasso type convex problem. In practice, just two iterations seem to be enough. A pseudocode for the ADMM algorithm used in this paper is given in Algorithm \ref{alg0} where we use the stopping (convergence) criterion following \cite[Sec.\ 3.3.1]{Boyd2004} and varying penalty parameter $\rho$ following \cite[Sec.\ 3.4.1]{Boyd2004}. The variables defined in (\ref{alg100})-(\ref{alg160}) are needed in Algorithm \ref{alg0} with ${\bm \Phi}_k^{(t+1)}$, ${\bm W}_k^{(t+1)}$, ${\bm U}_k^{(t+1)}$ as defined therein:
\begin{align}
& e_1 =  \|[{\bm \Phi}_1^{(t+1)}, \cdots , {\bm \Phi}_M^{(t+1)}]\|_F  \label{alg100} \\
& e_2 = \|[{\bm W}_1^{(t+1)}, \cdots , {\bm W}_M^{(t+1)}]\|_F \label{alg110} \\
& e_3 =  \|[{\bm U}_1^{(t+1)}, \cdots , {\bm U}_M^{(t+1)}]\|_F  \label{alg120} \\
& {\bm R}_p^{(t+1)} = \begin{bmatrix} {\bm \Phi}_1^{(t+1)} - {\bm W}_1^{(t+1)}, & \cdots , &  {\bm \Phi}_M^{(t+1)} - {\bm W}_M^{(t+1)}
           \end{bmatrix} \label{alg130} \\
& {\bm R}_d^{(t+1)} =  \rho^{(t)} \begin{bmatrix}  {\bm W}_1^{(t+1)} - {\bm W}_1^{(t)}, & \cdots , &
          {\bm W}_M^{(t+1)} - {\bm W}_M^{(t)} \end{bmatrix} \label{alg140} \\
&    \tau_{pri} =  mp \sqrt{M} \, \tau_{abs} + \tau_{rel} \, \max ( e_1, e_2 ) \label{alg150} \\
&   \tau_{dual} =  mp \sqrt{M} \, \tau_{abs} + \tau_{rel} \,  e_3 / \rho^{(t)} \, . \label{alg160}
\end{align}

Our ADMM-based optimization algorithm is as follows.
\begin{itemize}
\item[1.] Given $M$ and $K=2m_t+1$, calculate $\hat{\bm S}_k$. Initialize iteration $\tilde{m}=1$,  ${\bm \Omega}^{(0)} = {\bm 0}$, $\bar{\bm{\Omega}}= [\bar{\bm \Phi}_1, \cdots , \bar{\bm \Phi}_M] = {\bm \Omega}^{(0)}$ and use $\bar{\bm{\Omega}}$ to compute $\lambda_{kij}$'s and $\lambda_{q \ell M}$.
\item[2.] Execute Algorithm \ref{alg0} with initial guess $\bar{\bm \Phi}_k$, $k \in [M]$. 
\item[3.] Quit if using lasso, else set ${\bm \Omega}^{(\tilde{m})} = \hat{\bm \Omega}$ and $\bar{\bm{\Omega}} = {\bm \Omega}^{(\tilde{m})}$ to re-compute $\lambda_{kij}$'s and $\lambda_{q \ell M}$'s via the LLA. Let $\tilde{m} \leftarrow \tilde{m}+1$.
\item[4.] Repeat steps 2 and 3 until convergence. The converged $\hat{\bm \Omega}$ is the final estimate of the inverse PSD's. (For the numerical results shown in Sec.\ \ref{NE}, we terminated after two iterations of steps 2 and 3, similar to \cite{Zou2008, Lam2009}.)
\end{itemize}
For the numerical results in Sec.\ \ref{NE}, we used $\bar{\mu}=10$, $\bar{\rho}=2$, $\epsilon = 0.0001$ for log-sum penalty, $a$=3.7 (as in \cite{Fan2001, Lam2009}) for the SCAD penalty, $\tau_{abs} = \tau_{rel} = 10^{-4}$ and $t_{\max} = 200$.

\subsection{B.I.C.\ FOR TUNING PARAMETER SELECTION} \label{bic}
Given $n$ and choice of $K$ and $M$, we follow the Bayesian information criterion (BIC) as given in \cite{Tugnait22c}, to select $\lambda$ (with $\alpha=0.05$ fixed), for all penalty functions. The Bayesian information criterion (BIC) of \cite{Tugnait22c} is given by
\begin{align}
  {\rm BIC}&(\lambda , \alpha) =  2K  \sum_{k=1}^M \left( -\ln |\hat{\bm{\Phi}}_k|  + {\rm tr} \left( \hat{\bm S}_k  \hat{\bm{\Phi}}_k  \right) \right)  \nonumber \\
	&  + \ln (2 K M) \, \sum_{k=1}^M (\mbox{\# of nonzero elements in } \hat{\bm{\Phi}}_k ) 	  \, . \label{BIC}
\end{align}
We use BIC to select $\lambda$ from over a grid of values.  We search over $\lambda$ in the range $[\lambda_\ell , \lambda_u]$ selected via the following heuristic (similar to \cite{Tugnait21a, Tugnait22c}). We find the smallest $\lambda$, labeled $\lambda_{sm}$, for which we get a no-edge model (i.e., $| \hat{\cal E} | =0$). Then we set $\lambda_u = \lambda_{sm}/2$ and $\lambda_\ell = \lambda_u/10$. The given choice of $\lambda_u$ precludes ``extremely'' sparse models while that of $\lambda_\ell$ precludes ``very'' dense models. 

We note that there exist other general approaches for tuning parameter selection such as cross-validation. Cross-validation generally involves first partitioning the data into $K$-subsets ($K$=5 or 10 folds), i.e., $K$ non-overlapping subsets picked randomly. Then $K-1$ segments act as training data for model fitting and the remaining segment is used as test (or validation) set. Tuning parameter would be picked to minimize a test set measure (e.g., negative log-likelihood or some other non-penalized original objective function) after averaging over several partitions. For instance, \cite{Simon2013} uses such a method where the data is assumed to be i.i.d. For dependent data with frequency-domain approaches, there are several unresolved issues. For instance, the DFT over training and test datasets would have different resolution since sample size would be different. Moreover, to preserve time dependency, one cannot sample as for i.i.d.\ data; one must sample block-wise to keep contiguous data-points together. These are unresolved issues which precludes the use of cross-validation in our case.

\section{THEORETICAL ANALYSIS} \label{consist}
We now allow $p$, $M$, $K$ (see (\ref{window}), (\ref{windowM})), and $\lambda$ to be functions of sample size $n$, denoted as $p_n$, $M_n$, $K_n$ and $\lambda_n$, respectively. We take $p_n$ to be a non-decreasing function of $n$, as is typical in high-dimensional settings. Note that $K_n M_n \approx n/2$. Pick $K_n = a_1 n^\gamma$ and $M_n = a_2 n^{1-\gamma}$ for some $0.5 < \gamma < 1$, $0 < a_1, a_2 < \infty$, so that both $M_n/K_n \rightarrow 0$ and $K_n/n \rightarrow 0$ as $n \rightarrow \infty$ (cf.\ \cite[Remark 1]{Tugnait22c}). 

Recall that we have the original multi-attribute graph ${\cal G} = (V, {\cal E})$ with $|V|=p_n$ and the enlarged graph $\bar{\cal G} = (\bar{V}, \bar{\cal E})$ with $|\bar{V}|=mp_n$. We assume (A3) below regarding ${\cal G}$.
\begin{itemize}
\item[(A3)] Denote the true edge set of the graph by ${\cal E}_0$, implying that ${\cal E}_0 = \{ \{j,k\} ~:~ ({\bm S}^{-1}_{0}(f))^{(jk)} \not\equiv 0, ~j\ne k,  ~ 0 \le f \le 0.5 \}$ where ${\bm S}_0(f) $ denotes the true PSD of ${\bm x}(t)$. (We also use $\bm{\Phi}_{0k} $ for ${\bm S}^{-1}_{0}(\tilde{f}_k)$ where $\tilde{f}_k$ is as in (\ref{window}), and use ${\bm \Omega}_0$ to denote the true value of ${\bm \Omega}$). Assume that card$({\cal E}_0) =|{\cal E}_0| \le s_{n0}$.

\item[(A4)] The minimum and maximum eigenvalues of $mp_n \times mp_n$ PSD ${\bm S}_0(f)  \succ {\bm 0}$  satisfy 
\begin{align*}
     0 < \beta_{\min} & \le \min_{f \in [0,0.5]} \phi_{\min}({\bm S}_0(f)) \\
		    & \le 
		     \max_{f \in [0,0.5]} \phi_{\max}({\bm S}_0(f)) \le \beta_{\max} < \infty \, .
\end{align*}
Here $\beta_{\min}$ and $\beta_{\max}$ are not functions of $n$ (or $p_n$).
\end{itemize}

Let $\hat{\bm{\Omega}}_\lambda = \arg\min_{\bm{\Omega} \,:\, \bm{\Phi}_{k} \succ {\bm 0}}  \bar{\cal L}(\bm{\Omega})$.
Theorem 1  establishes local consistency of $\hat{\bm{\Omega}}_\lambda$ for non-convex penalties and global consistency for the convex penalty. \\
{\it THEOREM 1 (Local Consistency)}. For $\tau > 2$, let
\begin{equation}  \label{naeq58}
   C_0 = 80 \, \max_{\ell, f} ( [{\bm S}_0(f)]_{\ell \ell}) 
	    \sqrt{ N_0 / \ln ( m p_n )  }
\end{equation}
where 
\begin{equation}  \label{naeq58aa}
   N_0 = 2 \ln (16 (mp_n)^\tau M_n ) \, .
\end{equation}
Define
\begin{align}  
    R = & 8(1+m)C_0 / \beta_{\min}^2 \, , \label{neq15ab0} 
\end{align}
\begin{align}
    r_n = & \sqrt{ M_n(m p_n+ m^2 s_{n0}) \ln ( m p_n ) / K_n} = o(1)\, , \label{neq15ab1} 
\end{align}
\begin{align}
		N_1 = &  \arg \min \big\{ n \, : \, K_n > N_0 \big\} \, ,  \label{neq15ab2} 
\end{align}
\begin{align}
		N_2 = & \arg \min \left\{ n \, : \, r_n \le 
		     0.1/ \big( R \beta_{\min} \big) \right\} \, , \label{neq15ab3} 
\end{align}
\begin{align}
    N_3 = &  \arg \min \Big\{ n \, : \, r_n \le \epsilon/R \Big\} \, ,  \label{neq15ab4}  
\end{align}
\begin{align}
N_4 = &  \arg \min \Big\{n \, : \, \lambda_n \le 
		       \frac{\min_{(i,j): \, [{\bm \Omega}_{0}]_{ij} \ne 0} |[{\bm \Omega}_{0}]_{ij}|}{a+1} \Big\}  
					\, ,  \label{neq15ab4a}  
\end{align}
\begin{align}
		\lambda_{n \ell} = & 2 C_0  \sqrt{\ln (mp_n ) / K_n} \, ,  \label{neq15ab5} 
\end{align}
\begin{align}
		\lambda_{n u1} = & C_0 \, \frac{1+m}{m} \, 
		       \sqrt{ (m^2 + \frac{m p_n}{s_{n0}})\frac{\ln (mp_n )}{K_n}} \, ,  \label{neq15ab6} 
\end{align}
\begin{align}
		\lambda_{n u2} = & \max \left( R r_n , \lambda_{n u1} \right) \, .  \label{neq15ab7} 
\end{align}
Under assumptions (A1)-(A4), there exists a local minimizer $\hat{\bm{\Omega}}_\lambda$ of $\bar{\cal L}(\bm{\Omega})$ satisfying
\begin{equation}  \label{neq15}
  \| \hat{\bm{\Omega}}_\lambda - \bm{\Omega}_0 \|_F  \le R  r_n
\end{equation}
with probability $> \, 1-1/(mp_n)^{\tau-2}$ if
\begin{itemize}
\item[(i)] for the lasso penalty $\rho_\lambda(t) = \lambda |t|$,  sample size $n >  \max \{ N_1, N_2 \}$ and $\lambda_n$ satisfies $\lambda_{n \ell} \le \lambda_n \le \lambda_{nu1} \;$,
\item[(ii)] for the SCAD penalty $\rho_\lambda(t)$,  sample size $n >  \max \{ N_1, N_2, N_4 \}$ and  $\lambda_n = \lambda_{nu2} \;$,
\item[(iii)] sample size $n >  \max \{ N_1, N_2, N_3 \}$ and $\lambda_n$ satisfies $\lambda_{n \ell} \le \lambda_n \le \lambda_{nu1}$ for the log-sum penalty $\rho_\lambda(t)$.
\end{itemize}
For the lasso penalty, $\hat{\bm{\Omega}}_\lambda$ is a global minimizer whereas for the other two penalties, it is a local minimizer. $\quad \bullet$ \\
The proof of Theorem 1 is given in Appendix \ref{append1}. 

{\it REMARK 1}. Theorem 1 helps determine how to choose $M_n$ and $K_n$ so that for given $ p_n $, $\lim_{n \rightarrow \infty} \| \hat{\bm{\Omega}}_\lambda - \bm{\Omega}_0 \|_F = 0$ (see also \cite[Remark 2]{Tugnait22c}). This behavior is governed by  (\ref{neq15}), therefore we have to examine $r_n$. As noted before, since $K_n M_n \approx n/2$, if one picks $K_n = a_1 n^\gamma$, then $M_n = a_2 n^{1-\gamma}$ for some $0 < \gamma < 1$, $0 < a_1, a_2 < \infty$. Suppose that $p_n+ m s_{n0}$ satisfies $p_n+ m s_{n0} = a_3 n^\theta$ for some $0 \le \theta < 1$, $0 < a_3 < \infty$. Then for fixed $m$, we have
\begin{align}
  {\cal O} \left(  r_n \right) = & 
	 {\cal O} \left( \frac{(\ln(n))^{1/2} n^{(1-\gamma)/2} n^{\theta/2} }{n^{\gamma/2}} \right) \nonumber \\
	        = & {\cal O} \left(\frac{(\ln(n))^{1/2} }{ n^{(2 \gamma -1 -  \theta)/2} } \right) 
					 \overset{n \uparrow \infty}{\rightarrow}  0 
					 \mbox{ if } 2 \gamma -1 - \theta > 0 \, .  \label{neq1520}
\end{align}
Therefore, we must have $1 > \gamma > \frac{1}{2} + \frac{\theta}{2}$. If $\theta =0$ (fixed graph size and fixed number of connected edges w.r.t.\ sample size $n$), we need $\frac{1}{2} < \gamma < 1$. If $\theta > 0$, $\gamma$ has to be increased beyond what is needed for $\theta =0$, implying more smoothing of periodogram ${\bm d}_x(f_m) {\bm d}_x^H(f_m)$ around $f_k$ to estimate ${\bm S}_x(f_k)$ (recall (\ref{moresm})), leading to fewer frequency test points $M_n$. Clearly, we cannot have $\theta \ge 1$ because $p_n+ ms_{n0} = {\cal O}(n^\theta)$ will require $\gamma > 1$ which is impossible. 
$\quad \Box$

We follow the proof technique of \cite[Lemma 6]{Loh2017} in establishing  Lemma 1 whose proof is in Appendix \ref{append2}. \\
{\it LEMMA 1 (Local Convexity)}. The optimization problem
\begin{align}  
   & \hat{\bm{\Omega}}_\lambda =  \arg\min_{ \bm{\Omega} \,:\, \bm{\Phi}_{k} \in {\cal B}_k }  
		\bar{\cal L}(\bm{\Omega}) \, , \label{neq4000} \\
	&	{\cal B}_k =  \Big\{ \bm{\Phi}_{k} \,:\, \bm{\Phi}_{k} \succ {\bm 0} , \; 
		   \| \bm{\Phi}_{k} \| \le 0.99 \, \sqrt{2/(m \mu \sqrt{M_n} \,)} \, \Big\} \, , \\
	&	\sqrt{2/(m \mu \sqrt{M_n} \,)} =  \left\{ \begin{array}{ll}
		   \infty & : \;\; \mbox{Lasso} \\
			  \sqrt{\frac{2(a-1)}{m \sqrt{M_n}}} & : \;\; \mbox{SCAD} \\
				\sqrt{ \frac{2 \epsilon}{m \sqrt{M_n} \lambda_n} } & : \;\; \mbox{log-sum}, \end{array} \right. \label{neq4010} 
\end{align}
consists of a strictly convex objective function over a convex constraint set, for all three penalties, where $\lambda_n$ is as defined in Theorem 1. $\quad \bullet$

Lemma 1 and Theorem 1 lead to Theorem 2 which is proved in Appendix \ref{append2}. \\
{\it THEOREM 2}. Assume the conditions of Theorem 1. Then $\hat{\bm{\Omega}}_\lambda$ as defined in Lemma 1 is unique, satisfying $\| \hat{\bm{\Omega}}_\lambda - \bm{\Omega}_0 \|_F  \le R  r_n$ with probability $> \; 1-1/(mp_n)^{\tau-2}$ if $R r_n + 1/\beta_{\min} \le 0.99 \, \sqrt{2/(m \mu \sqrt{M_n} \,)}$, as defined in Lemma 1. $\quad \bullet$ 

{\it REMARK 2}. With lasso, (\ref{neq4000}) is obviously a globally convex optimization problem since ${\cal B}_k =  \Big\{ \bm{\Phi}_{k} \,:\, \bm{\Phi}_{k} \succ {\bm 0} \Big\}$, hence, Theorems 1 and 2 yield a unique global optimum. For the SCAD penalty, $ \sqrt{\frac{2(a-1)}{m \sqrt{M_n}}} = {\cal O}(a^{1/2}/n^{(1-\gamma)/4})$ with $M_n$ as in Remark 1. For fixed SCAD parameter $a$, with increasing $n$ the convexity region shrinks. To counter this, one could allow $a$ to increase, but this would make SCAD more like lasso. To consider log-sum penalty, using (\ref{neq15ab1}) and (\ref{neq15ab6}), we express $\lambda_{n u1}$ as $\lambda_{n u1} = C_0 (1+m) r_n / \sqrt{m^2 s_{n0} M_n}$, which together with $\lambda_n \le \lambda_{nu1}$  implies that
\begin{align}
  \sqrt{ \frac{2 \epsilon}{m \sqrt{M_n} \lambda_n} } \; \ge & \;
	 \sqrt{ \frac{2 \epsilon \sqrt{s_{n0}}}{C_0 (1+m) r_n  } } 
%	        =  {\cal O} \left(r_n^{-1/2} \right) 
					 \overset{n \uparrow \infty}{\rightarrow}  \, \infty \, .  \label{neq1520b}
\end{align}
Now with increasing $n$, the convexity region expands, unlike SCAD. 
$\quad \Box$

We now turn to graph recovery. We follow the proof technique of \cite[Theorem 10]{Zhao2022} in establishing  Theorem 3 whose proof is in Appendix \ref{append2}. For some $\gamma_n >0$, define
\begin{align}  
    \hat{\cal E} = & \left\{ \{q,\ell\} \, : \, \|\hat{\bm \Omega}^{(q \ell M_n)} \|_F > \gamma_n > 0,
		    q \ne \ell \right\} \, , \label{neq4020} \\
		{\cal E}_0 = &   \left\{ \{q,\ell\} \, : \, \|{\bm \Omega}_0^{(q \ell M_n)} \|_F >  0,
		    q \ne \ell \right\} \, , \label{neq4022} \\
       \bar{\sigma}_n = &   R r_n   \, ,  \label{neq4022} \\
%			\;\; \nu  =   \min_{\{q,\ell\} \in {\cal E}_0} \,  \|{\bm \Omega}_0^{(q \ell M_n)} \|_F  \, , \label{neq4024} \\
		\nu  = &  \min_{\{q,\ell\} \in {\cal E}_0} \,  \|{\bm \Omega}_0^{(q \ell M_n)} \|_F  \, , \label{neq4024} \\
			N_4 = & \arg \min \Big\{ n \, : \, \bar{\sigma}_n \le 0.4 \nu \Big\} \, , \label{neq4025}
\end{align}
where $R$ and $r_n$ are as in (\ref{neq15ab0}) and (\ref{neq15ab1}), respectively. \\
{\it THEOREM 3}. For $\gamma_n = 0.5 \nu$ and $n \ge N_4$, $\hat{\cal E} = {\cal E}_0$ with probability $> \; 1-1/(mp_n)^{\tau-2}$ under the conditions of Theorem 1.  $\quad \bullet$ 

{\it REMARK 3}.  In practice we do not know the value of $\nu$, hence cannot calculate $\gamma_n$ needed in (\ref{neq4020}). For the numerical results presented in Sec.\ \ref{NE}, we used $\gamma_n =0$. Using some incoherence or irrepresentability conditions and the primal-dual witness method (as in \cite{Ravikumar2011, Kolar2014}), it may be possible to establish a result similar to Theorem 3 but with $\gamma_n =0$. We leave this for future research. We do not impose any incoherence or irrepresentability conditions in this paper.
$\quad \Box$

{\it REMARK 4}.  We now provide a detailed comparison between this paper and \cite{Tugnait22c} (also \cite{Tugnait22d}).  The differences between this paper and \cite{Tugnait22c, Tugnait22d} are as follows.
\begin{itemize}
\item[(i)] As discussed in Secs.\ \ref{SM} and \ref{opt}, in this paper we have $(mp) \times (mp)$ inverse PSD matrices ${\bm \Phi}_k$, $k \in [M]$, for a $p$-node graph, compared to $p \times p$ ${\bm \Phi}_k$'s in \cite{Tugnait22c, Tugnait22d}. This requires larger groups comprised of $M m^2$ variables in the group penalty term $P_g({\bm \Omega})$ given by (\ref{eqth2_20b}), compared to groups of $M$ variables in \cite{Tugnait22c, Tugnait22d}. In \cite{Tugnait22c, Tugnait22d} the group penalty term is missing the factor $\sqrt{M}$ (see \cite[(41)]{Tugnait22c}) corresponding to the factor $m \sqrt{M}$ in (\ref{eqth2_20b}) of this paper. This factor equals the square-root of the number of group variables, following the work of \cite{Yuan2007}. A consequence of the extra factor $m \sqrt{M}$ is that in Theorem 1 of this paper, the bounds on $\lambda_n$ for the lasso penalty do not depend on $\alpha$ (see (\ref{neq15ab5}) and (\ref{neq15ab6})), whereas the corresponding result (with $m=1$) in \cite[(69)]{Tugnait22c} depends upon $\alpha$. In \cite[Theorem 1]{Tugnait22c} the lower bound on $\lambda_n$ can be greater than the upperbound for certain choices of a constant $C_1$ whereas no such anomaly arises in this paper.

\item[(ii)] In \cite{Tugnait22c} non-convex penalties are not considered. In \cite{Tugnait22d} non-convex log-sum regularization for CIG learning for single-attribute Gaussian time series has been proposed replacing the lasso penalty of \cite{Tugnait22c}. It is shown empirically in \cite{Tugnait22d} that the log-sum penalty significantly outperforms the lasso penalty with $F_1$ score as a performance measure. A theorem corresponding to Theorem 1 of 
this paper and that of \cite{Tugnait22c} is stated in \cite{Tugnait22d} without any proof. Moreover, as in \cite{Tugnait22c}, for lack of the factor $\sqrt{M}$ in the group penalty term in \cite{Tugnait22d}, the upperbound on $\lambda_n$ in \cite[Theorem]{Tugnait22d} depends on $\alpha$ and it can be smaller than the lowerbound for certain choices of a constant. No such anomaly arises in this paper.

\item[(iii)] In this paper we provide results for two non-convex penalties (SCAD and log-sum) for CIG learning from multi-attribute time series. The two penalties require different analysis in proving Theorem 1 (compare (\ref{expan})-(\ref{Armain}) for lasso and log-sum penalties with (\ref{expan2})-(\ref{naeq92152}) for the SCAD penalty in Appendix \ref{append1}). We provide a complete proof of Theorem 1 whereas \cite{Tugnait22d} has no proof of its theorem.

\item[(iv)] There are no results corresponding to our Lemma 1 and Theorems 2 and 3 in \cite{Tugnait22c, Tugnait22d}  (not needed in \cite{Tugnait22c} since it does not consider non-convex penalties). $\quad \Box$
\end{itemize}

\begin{figure*}
\begin{subfigure}[b]{0.5\textwidth}
\begin{center}
\includegraphics[width=0.8\linewidth]{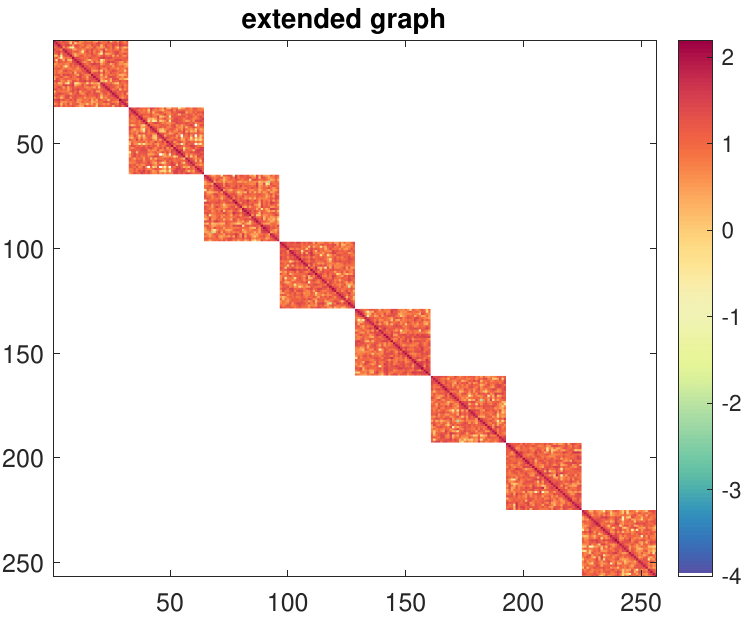}
\caption{Model  1}
\end{center}
\end{subfigure}%
\begin{subfigure}[b]{0.5\textwidth}
\begin{center}
\includegraphics[width=0.8\linewidth]{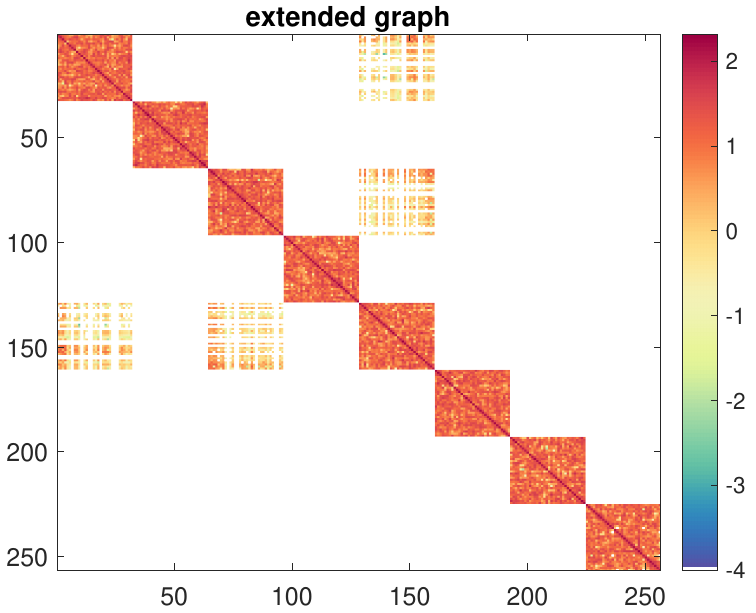}
\caption{Model  2}
\end{center}
\end{subfigure}%
\caption{\small{ True $\log_{10} \big(\sum_{f=0:0.01:5} | [S^{-1}(f)]_{ij} | \big)$, $i,j \in [256]$, for extended graphs for a single Monte Carlo run: $mp=4 \times 64 = 256$ nodes.  }} \label{figsyn}
\end{figure*}

\section{NUMERICAL EXAMPLES}  \label{NE}
In this section we present numerical results using both synthetic and real data to illustrate the proposed approach. We know the ground truth in the synthetic data example which permits assessment of the efficacy of our approaches. The ground truth is unknown in the real data example and here we wish to visualize and explore the conditional dependency structure underlying the data.

\subsection{SYNTHETIC DATA} Consider a graph with $p=64$ nodes, each node with $m=4$ attributes. The time series data $\{ {\bm x}(t) \}$ is generated using a vector autoregressive model of order 3 (VAR(3)): 
\begin{equation}  \label{VAR1}
    {\bm x}(t) = \sum_{i=1}^3 {\bm A}_i {\bm x}(t-i) + {\bm w}(t) \, , \quad {\bm x}(t) \in \mathbb{R}^{mp} \, ,
\end{equation} 
where ${\bm w}(t)$ is i.i.d.\ zero-mean Gaussian with precision matrix either $\tilde{\bm \Omega}= \tilde{\bm \Omega}_1$  (labeled Model  1) or $\tilde{\bm \Omega}= \tilde{\bm \Omega}_1 + \tilde{\bm \Omega}_2$  (labeled Model  2). For Model  1, we create 8 clusters (communities) of 8 nodes each, each node with $m=4$ attributes, where nodes within a community are not connected to any node in other communities. To generate ${\tilde{\bm \Omega}}_1$,  we set $[{\tilde{\bm \Omega}}_1^{(q \ell)}]_{uv} = 0.5^{|u-v|}$ for $q = \ell \in [8]$, $u \ne v, \; u,v \in [m]$ (notation as in \ref{neweq12}), and it is zero otherwise. For $q \ne \ell$, we have  ${\tilde{\bm \Omega}}_1^{(q \ell)} = {\bm 0}$. We add $\gamma {\bm I}_{mp}$ to $\tilde{\bm \Omega}_1$ and choose $\gamma$ to make the minimum eigenvalue of $\tilde{\bm \Omega}_1+\gamma {\bm I}_{mp}$ equal to 0.5$~$. The parameters of VAR(3) model are generated similarly by having $\bm{A}_i^{(q \ell)} = {\bm 0}$ for $q \ne \ell$, and only 10\% of the entries of $\bm{A}_i^{(q q)}$'s are nonzero with the nonzero elements independently and uniformly distributed over $[-0.6,0.6]$. We then check if the VAR(3) model is stable, a necessary and sufficient condition for which is that the roots of $a(z) = |{\bm I}_{mp}-\sum_{i=1}^3 {\bm A}_i z^{-i}| = 0$ should all have modulus $< 1$; this condition is equivalent to having all eigenvalues of the corresponding $(3mp)\times (3mp)$ companion matrix to have modulus $<1$ \cite[Sec.\ 8.2.3]{Tsay2010}. Additionally, in order to avoid a ``long'' impulse response, we require the roots of $a(z)$ to have modulus $\le 0.95$. Suppose this condition is violated with $|z_{\max}| > 0.95$ where $|z_{\max}| = \arg\max_{\ell \in [3mp]} \{ |z_\ell| \, :\, a(z_\ell) = 0\}$.  In this case, we scale ${\bm A}_i$'s to $\bar{\bm A}_i = \gamma^i {\bm A}_i$, $\gamma = 0.95/|z_{\max}|$. It is easy to see that the roots of $\bar{a}(z) = |{\bm I}_{mp}-\sum_{i=1}^3 \bar{\bm A}_i z^{-i}| =a(z/\gamma)= 0$ now all have modulus $\le 0.95$. 

For Model  2, we allow some interaction between the 8 clusters via $\tilde{\bm \Omega}_2$ which is generated via an Erd\"{o}s-R\`{e}nyi graph structure where the $p$ nodes are connected with probability $p_{er} =0.002$. To generate $\tilde{\bm \Omega}_2$,  we set ${\tilde{\bm \Omega}}_2^{(q \ell)} = {\bm 0}$ for $q = \ell \in [8]$, and for $q \ne \ell$ but connected in the Erd\"{o}s-R\`{e}nyi graph, the entries of  ${\tilde{\bm \Omega}}_2^{(q \ell)}$ are independently and uniformly distributed over $[-0.4,-0.1] \cup [0.1,0.4]$,  and are zero if not connected.

First 100 samples are discarded to eliminate transients. This set-up leads to approximately 11\% and 13\% connected edges in models 1 and 2, respectively. In each run, we calculated the true PSD ${\bm S}(f)$ for $f \in [0,0.5]$ at intervals of 0.01, and then take $\{q,\ell\} \in {\cal E}$ if $\sqrt{\sum_f \|( {\bm S}^{-1}(f))^{(q \ell)}\|_F^2 }> 10^{-2} (\max_{q, \ell \in [p]} \sqrt{\sum_f \|( {\bm S}^{-1}(f))^{(q \ell)}\|_F^2 }) $, else $\{q,\ell\} \not\in {\cal E}$. For a typical realization (run), Fig.\ \ref{figsyn} shows heatmaps of $\log_{10} \big(\sum_{f=0:0.01:5} | [S^{-1}(f)]_{ij} | \big)$, $i,j \in [256]$, for models 1 and 2.
\begin{table}
\caption{\small{\it Model  1: $F_1$ scores, Hamming distances and timings, averaged over 100 runs.}} \label{table1} 
\begin{center}
\begin{tabular}{cccc}   \hline\hline
 $n$ &  128 &  256  & 1024  \\  \hline\hline
\multicolumn{4}{c}{$M$=4: $F_1$ score $\pm \sigma$: $\lambda$'s picked to maximize $F_1$ } \\ \hline
Lasso  &  0.5788 $\pm$ 0.1407  &  0.7647 $\pm$ 0.1308  & 0.9682 $\pm$ 0.0347     \\  
Log-sum   &  0.7065 $\pm$ 0.0517  &  0.8679 $\pm$ 0.0261  & 0.9899 $\pm$ 0.0077     \\ 
SCAD   &  0.5820 $\pm$ 0.1428  &  0.7651 $\pm$ 0.1312  & 0.9675 $\pm$ 0.0347     \\ \hline\hline
\multicolumn{4}{c}{$M$=4: Hamming distance $\pm \sigma$: $\lambda$'s picked to maximize $F_1$ } \\ \hline
Lasso  &  168.53 $\pm$ 040.255  &  097.36 $\pm$ 044.03  & 013.93 $\pm$ 014.71     \\  
Log-sum   &  113.32 $\pm$ 012.37  &  057.70 $\pm$ 011.05  & 004.46 $\pm$ 003.34     \\ 
SCAD   &  165.41 $\pm$ 037.59  &  097.14 $\pm$ 044.15  & 014.19 $\pm$ 014.66     \\ \hline\hline
\multicolumn{4}{c}{$M$=4: Timing (s) $\pm \sigma$: $\lambda$'s picked to maximize $F_1$ } \\ \hline
Lasso  &  011.45 $\pm$ 01.105  &  009.52 $\pm$ 01.477  & 005.65 $\pm$ 00.585     \\  
Log-sum   &  019.62 $\pm$ 00.497  &  016.88 $\pm$ 01.309  & 010.94 $\pm$ 00.808     \\ 
SCAD   &  023.79 $\pm$ 02.005  &  019.29 $\pm$ 02.774  & 011.92 $\pm$ 01.206     \\ \hline\hline
\multicolumn{4}{c}{$M$=4: $F_1$ score $\pm \sigma$: $\lambda$'s picked to minimize BIC } \\ \hline
Log-sum   &  0.4394 $\pm$ 0.0106  &  0.6632 $\pm$ 0.0496  & 0.9577 $\pm$ 0.0534     \\ \hline\hline
\multicolumn{4}{c}{$M$=4: Hamming distance $\pm \sigma$: $\lambda$'s picked to minimize BIC } \\ \hline
Log-sum   &  499.97 $\pm$ 015.93  &  214.10 $\pm$ 050.73  & 017.22 $\pm$ 020.15     \\ \hline\hline
\end{tabular} 
\end{center}
\end{table}

\begin{table}
\caption{\small{\it Model  2: $F_1$ scores, Hamming distances and timings, averaged over 100 runs.}} \label{table2} 
\begin{center}
\begin{tabular}{cccc}   \hline\hline
 $n$ &  128 &  256  & 1024  \\  \hline\hline
\multicolumn{4}{c}{$M$=4: $F_1$ score $\pm \sigma$: $\lambda$'s picked to maximize $F_1$ } \\ \hline
Lasso  &  0.4907 $\pm$ 0.0853  &  0.6098 $\pm$ 0.1460  & 0.7847 $\pm$ 0.0911     \\  
Log-sum   &  0.5692 $\pm$ 0.0346  &  0.7241 $\pm$ 0.0599  & 0.8236 $\pm$ 0.0696     \\ 
SCAD   &  0.4982 $\pm$ 0.0862  &  0.6087 $\pm$ 0.1456  & 0.7819 $\pm$ 0.0911     \\ \hline\hline
\multicolumn{4}{c}{$M$=4: Hamming distance $\pm \sigma$: $\lambda$'s picked to maximize $F_1$ } \\ \hline
Lasso  &  307.37 $\pm$ 115.32  &  219.54 $\pm$ 147.61  & 126.68 $\pm$ 103.50     \\  
Log-sum   &  241.40 $\pm$ 038.08  &  145.72 $\pm$ 043.82  & 099.39 $\pm$ 047.80     \\ 
SCAD   &  387.51 $\pm$ 115.59  &  219.77 $\pm$ 147.05  & 127.69 $\pm$ 102.12     \\ \hline\hline
\multicolumn{4}{c}{$M$=4: Timing (s) $\pm \sigma$: $\lambda$'s picked to maximize $F_1$ } \\ \hline
Lasso  &  10.598 $\pm$ 01.167  &  09.090 $\pm$ 01.534  & 06.790 $\pm$ 01.190     \\  
Log-sum   &  20.141 $\pm$ 00.719  &  16.882 $\pm$ 01.519  & 12.783 $\pm$ 01.526     \\ 
SCAD   &  20.555 $\pm$ 02.474  &  17.446 $\pm$ 02.908  & 12.562 $\pm$ 01.831     \\ \hline\hline
\multicolumn{4}{c}{$M$=4: $F_1$ score $\pm \sigma$: $\lambda$'s picked to minimize BIC } \\ \hline
Log-sum   &  0.5136 $\pm$ 0.0195  &  0.7224 $\pm$ 0.0644  & 0.7623 $\pm$ 0.1030     \\ \hline\hline
\multicolumn{4}{c}{$M$=4: Hamming distance $\pm \sigma$: $\lambda$'s picked to minimize BIC } \\ \hline
Log-sum   &  359.59 $\pm$ 039.91  &  139.84 $\pm$ 039.46  & 115.90 $\pm$ 050.71     \\ \hline\hline
\end{tabular} 
\end{center}
\end{table}

\begin{table}
\caption{\small{\it Model  2: $F_1$ scores and  Hamming distances using log-sum penalty, averaged over 100 runs.}} \label{table3} 
\begin{center}
\begin{tabular}{cccc}   \hline\hline
 $n$ &  128 &  256  & 1024  \\  \hline\hline
\multicolumn{4}{c}{$F_1$ score $\pm \sigma$: $\lambda$'s picked to maximize $F_1$ } \\ \hline
$M$=2  &  0.5967 $\pm$ 0.0440  &  0.7324 $\pm$ 0.0692  & 0.8283 $\pm$ 0.0695     \\  
$M$=3   &  0.5826 $\pm$ 0.0396  &  0.7294 $\pm$ 0.0664  & 0.8269 $\pm$ 0.0691     \\ 
$M$=4   &   0.5692 $\pm$ 0.0346  &  0.7241 $\pm$ 0.0599  & 0.8236 $\pm$ 0.0696     \\ 
$M$=6   &  0.5156 $\pm$ 0.0358  &  0.7046 $\pm$ 0.0611  & 0.8158 $\pm$ 0.0716     \\ \hline\hline
\multicolumn{4}{c}{Hamming distance $\pm \sigma$: $\lambda$'s picked to maximize $F_1$ } \\ \hline
$M$=2    &  215.13 $\pm$ 039.42  &  134.40 $\pm$ 039.84  & 096.64 $\pm$ 047.14     \\  
$M$=3   &  224.72 $\pm$ 037.49  &  136.61 $\pm$ 040.49  & 097.85 $\pm$ 047.42     \\ 
$M$=4   &  241.40 $\pm$ 038.08  &  145.72 $\pm$ 043.82  & 099.39 $\pm$ 047.80     \\
$M$=6   &  261.21 $\pm$ 036.99  &  149.50 $\pm$ 038.84  & 104.11 $\pm$ 051.47    \\ \hline\hline
\end{tabular} 
\end{center}
\end{table}

\begin{table}
\caption{\small{\it Model  2, varying AR model order: VAR($L$) as in (\ref{VARL}), $L \in \{1,2,3,4 \}$. $F_1$ scores and  Hamming distances using log-sum penalty, averaged over 100 runs.}} \label{table4} 
\begin{center}
\begin{tabular}{cccc}   \hline\hline
 $n$ &  128 &  256  & 1024  \\  \hline\hline
\multicolumn{4}{c}{$M$=4: $F_1$ score $\pm \sigma$: $\lambda$'s picked to maximize $F_1$ } \\ \hline
$L$=1  &  0.5045 $\pm$ 0.0273  &  0.6584 $\pm$ 0.0522  & 0.8067 $\pm$ 0.0500     \\  
$L$=2   &  0.5834 $\pm$ 0.0726  &  0.7050 $\pm$ 0.0622  & 0.8212 $\pm$ 0.0703     \\ 
$L$=3   &   0.5692 $\pm$ 0.0346  &  0.7241 $\pm$ 0.0599  & 0.8236 $\pm$ 0.0696     \\ 
$L$=4   &  0.5609 $\pm$ 0.0476  &  0.7191 $\pm$ 0.0625  & 0.8361 $\pm$ 0.0632     \\ \hline\hline
\multicolumn{4}{c}{$M$=4: Hamming distance $\pm \sigma$: $\lambda$'s picked to maximize $F_1$ } \\ \hline
$L$=1    &  258.80 $\pm$ 028.79  &  161.73 $\pm$ 046.36  & 097.67 $\pm$ 032.57     \\  
$L$=2   &  202.80 $\pm$ 046.89  &  156.61 $\pm$ 052.88  & 098.76 $\pm$ 046.74     \\ 
$L$=3   &  241.40 $\pm$ 038.08  &  145.72 $\pm$ 043.82  & 099.39 $\pm$ 047.80     \\
$L$=4   &  248.40 $\pm$ 046.51  &  149.73 $\pm$ 046.00  & 090.37 $\pm$ 041.32    \\ \hline\hline
\end{tabular} 
\end{center}
\end{table}

Simulation results based on 100 runs are shown in Tables \ref{table1}-\ref{table4} where the performance measures are $F_1$-score and Hamming distance for efficacy in edge detection. All algorithms were run on a Window 10 Pro operating system with processor Intel(R) Core(TM) i7-10700 CPU @2.90 GHz with 32 GB RAM, using MATLAB R2023a. The $F_1$-score is defined as $F_1 = 2 \times \mbox{precision} \times \mbox{recall}/(\mbox{precision} + \mbox{recall})$ where $\mbox{precision} = | \hat{\cal E} \cap {\cal E}_0|/ |\hat{\cal E}|$, $\mbox{recall} = |\hat{\cal E} \cap {\cal E}_0|/ |{\cal E}_0|$, and ${\cal E}_0$ and $ \hat{\cal E}$ denote the true and estimated edge sets, respectively. The Hamming distance is between $\hat{\cal E} $ and ${\cal E}_0$, scaled by 0.5 to count only distinct edges. For our proposed approach, we consider $M=4$ for three samples sizes $n \in \{128, 256, 1024 \}$ in Table \ref{table1} for Model 1 and Table \ref{table2} for Model 2. For  $M=4$, we used $K=2m_t+1=15,31,127$ for $n=128,256,1024$, respectively. We fixed $\alpha=0.05$ and $\lambda$ was selected by searching over a grid of values to maximize the $F_1$-score (over 100 runs), or via BIC as in Sec.\ \ref{opt}-\ref{bic}. We used lasso (convex), log-sum (non-convex, $\epsilon = 0.0001$) or SCAD (non-convex, $a$=3.7) penalties. When $\lambda$'s are picked to maximize the $F_1$ score, it is seen that the log-sum penalty outperforms the lasso and the SCAD penalties in both Table \ref{table1} (Model 1) and Table \ref{table2} (Model 2) in terms of the $F_1$-score as well as the Hamming distance, whereas the SCAD penalty does not offer much improvement over lasso. As discussed in Remark 2, the ``convexity'' region for the log-sum penalty is likely to be much larger than that for SCAD.  With the lasso penalty, computational time is close to half of that for log-sum or SCAD, which is not surprising since the latter are initialized using the lasso result (cf.\ Sec.\ \ref{opt}). When $\lambda$'s are picked via BIC (only for the log-sum penalty), there is a drop in the $F_1$ score and increase in the Hamming distance as compared to the case where $\lambda$'s are picked to maximize the $F_1$ score. This is due to errors in the BIC parameter selection method.

In Table \ref{table3} we show the results for the log-sum penalty for $M=2,3,4$ and 6 with $\lambda$ selected to maximize the $F_1$-score. We take $n=128, 256, 1024$ and the corresponding  $m_t$ values leading to different $M$ values are $m_t= 15, 31, 127 $ ($M=2$), $m_t= 9, 20, 84 $ ($M=3$), $m_t= 7, 15, 63$ ($M=4$), and $m_t= 4, 10, 42$ ($M=6$). The number of unknown parameters being estimated are ${\cal O}(M (mp)^2)$ for $M$ $(mp) \times (mp)$ ${\bm \Phi}_k$'s. We see that for a fixed $n$, at first the performance changes only a little with increasing $M$, then it declines more sharply ($M=$ 4 to 6) as more parameters need to be estimated with increasing $M$.

In Table \ref{table4} we display some numerical ablation results by varying the AR model order. We use a VAR($L$) model with $L \in \{1,2,3,4 \}$, given by 
\begin{equation}  \label{VARL}
    {\bm x}(t) = \sum_{i=1}^L {\bm A}_i {\bm x}(t-i) + {\bm w}(t) \, , \quad {\bm x}(t) \in \mathbb{R}^{mp} \, ,
\end{equation}  
where ${\bm A}_i$'s are picked as for (\ref{VAR1}) and we used Model 2 to specify the precision matrix of ${\bm w}(t)$. We used the log-sum penalty and $M=4$. It is seen that the results are consistent across model orders, both in terms of the $F_1$ scores and the Hamming distances.

\begin{figure*}
\begin{subfigure}[b]{0.5\textwidth}
\begin{center}
\includegraphics[width=0.8\linewidth]{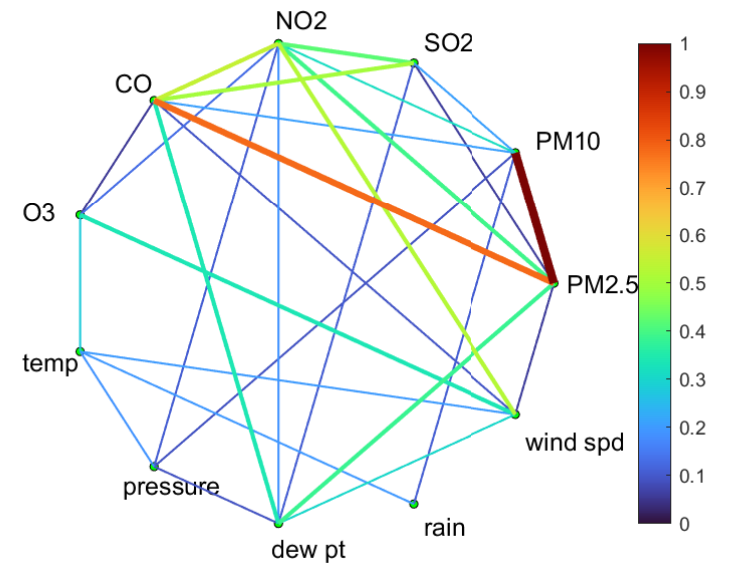}
\caption{M=4: Lasso penalty}
\end{center}
\end{subfigure}%
\begin{subfigure}[b]{0.5\textwidth}
\begin{center}
\includegraphics[width=0.8\linewidth]{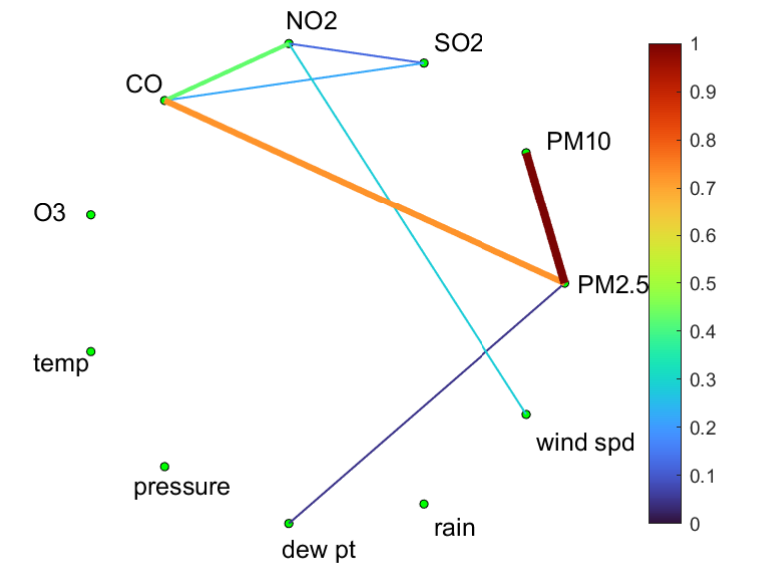}
\caption{M=4: Log-sum penalty}
\end{center}
\end{subfigure}%
\caption{\small{Pollution graphs for the Beijing air-quality dataset \cite{Zhang2017} for year 2013-14: 8 monitoring sites and 11 features ($m=8$, $p=11$, $M=4$, $n=364$). Number of distinct edges $=29$ and $7$ in graphs (a) and (b), respectively. Estimated $\|\hat{\bm \Omega}^{(ijM)}\|_F$ is the edge weight (normalized to have $\max_{i \ne j}\|\hat{\bm \Omega}^{(ijM)}\|_F=1$), see (\ref{eqth2_20c}). The edge weights are color coded , in addition to the edges with higher weights being drawn thicker.}} \label{figreal}
\vspace*{-0.2in}
\end{figure*}
%\vspace*{-0.2in}
\begin{figure*}
\begin{subfigure}[b]{0.5\textwidth}
\begin{center}
\includegraphics[width=0.8\linewidth]{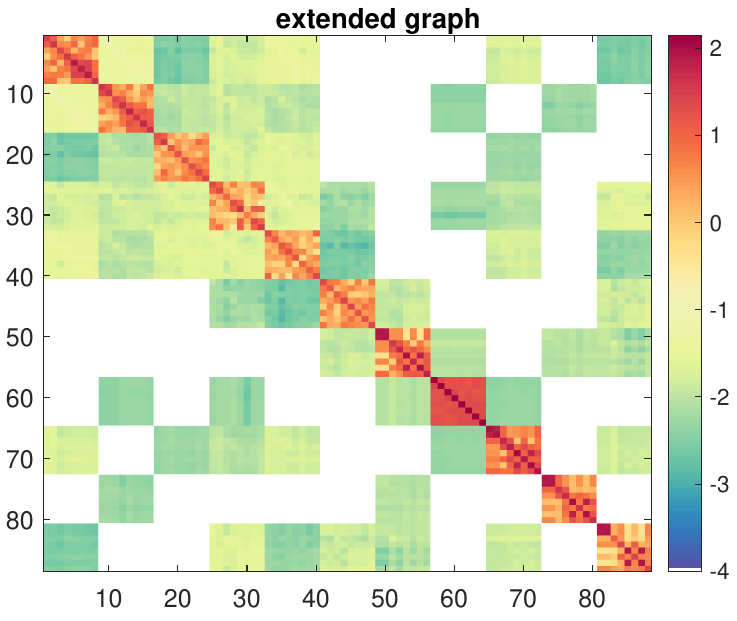}
\caption{Lasso}
\end{center}
\end{subfigure}%
\begin{subfigure}[b]{0.5\textwidth}
\begin{center}
\includegraphics[width=0.8\linewidth]{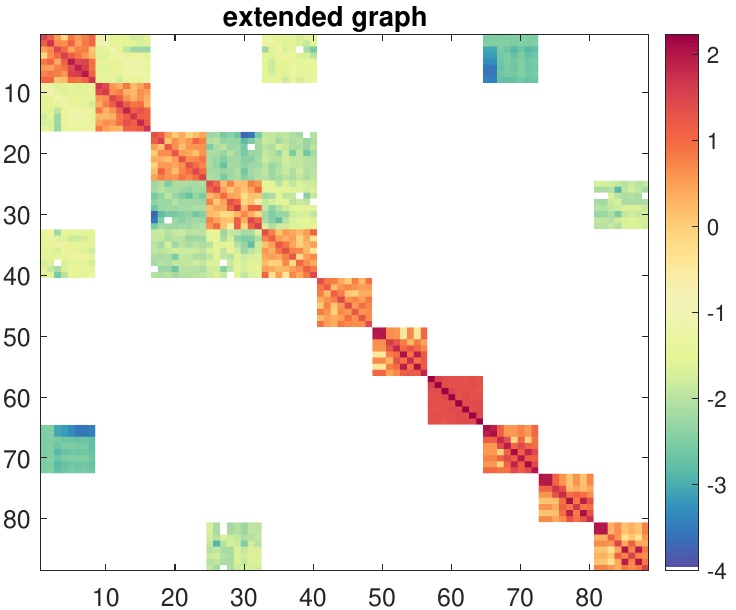}
\caption{Log-sum}
\end{center}
\end{subfigure}%
\caption{\small{ Estimated $\log_{10} \big( \sqrt{\sum_{k=1}^M | [\hat{\bm \Phi}_k]_{ij} |^2 } \big)$, $i,j \in [88]$, for the Beijing air-quality dataset ($m=8$, $p=11$, $M=4$, $n=364$). There are $p=11$ nodes (PM$_{2.5}$ labeled as node 1, PM$_{10}$ as 2, and so on, moving counter-clockwise in Fig.\ \ref{figreal}), each variables measured at $m=8$ stations.}} \label{figreal2}
\end{figure*}

\subsection{REAL DATA: BEIJING AIR-QUALITY DATASET \cite{Zhang2017}} \label{NEreal}
Here we consider Beijing air-quality dataset \cite{Zhang2017, Chen2015}, downloaded from \url{https://archive.ics.uci.edu/dataset/501/beijing+multi+site+air+quality+data}. This data set includes hourly air pollutants data from 12 nationally-controlled air-quality monitoring sites in the Beijing area. The time period is from March 1st, 2013 to February 28th, 2017. The six air pollutants are PM$_{2.5}$, PM$_{10}$, SO$_2$, NO$_2$, CO, and O$_3$, and the meteorological data is comprised of five features: temperature, atmospheric pressure, dew point, wind speed, and rain; we did not use wind direction. Thus we have eleven ($=p$) features (pollutants and weather variables). We used data from 8 ($=m$) sites:  Changping, Dingling, Huairou, Shunyi, Aotizhongxin, Dongsi, Guanyuan, Gucheng. The data are averaged over 24 hour period to yield daily averages $x_i(t)$, $i \in [88]$. We used one year 2013-14 of daily data resulting in $n = 365$ days. 
We pre-processed the data as follows. Given $x_i(t)$, we transform it to $\bar{x}_i(t) = \ln(x_i(t)/x_i(t-1))$ for each $i$ (leads to $n=364$), and then detrend it (i.e., remove the best straight-line fit). Finally, we scale the detrended scalar sequence to have a mean-square value of one. All temperatures were converted from Celsius to Kelvin to avoid negative numbers. If a value of a feature is zero (e.g., wind speed), we added a small positive number to it so that the log transformation is well-defined.  

Fig.\ \ref{figreal} shows the CIGs for lasso and log-sum penalties for $M=4$ where with $\alpha=0.05$, $\lambda$ was selected via BIC: an edges exists iff $\|\hat{\bm \Omega}^{(ijM)}\|_F > 0$. 
The corresponding heatmaps showing estimated $\log_{10} \big( \sqrt{\sum_{k=1}^M | [\hat{\bm \Phi}_k]_{ij} |^2 } \big)$, $i,j \in [88]$ are in Fig.\ \ref{figreal2}.  
It is seen that lasso yields a much denser graph (29 edges) while the graph resulting from the log-sum penalty is much sparser (7 edges). Cold, dry air from the north of Beijing reduces both dew point and PM$_{2.5}$ particle concentration in suburban areas while southerly wind brings warmer and more humid air from the more polluted south that elevates both dew point and PM$_{2.5}$ concentration \cite{Zhang2017}. This fact is captured by the edge between dew point and PM$_{2.5}$ in Fig.\ \ref{figreal}.

\section{CONCLUSIONS} Estimation of the CIG of high-dimensional multivariate Gaussian time series from multi-attribute data was considered. We provided a unified theoretical analysis of multi-attribute graph learning for dependent time series using a penalized log-likelihood objective function in the frequency-domain. Both convex and non-convex regularization functions were considered. We established  sufficient conditions for consistency, local convexity when using non-convex penalties, and graph recovery. Our approach was illustrated using numerical examples utilizing both synthetic and real (Beijing air-quality dataset) data. Non-convex log-sum regularization yielded more accurate results compared to convex sparse-group lasso  or non-convex SCAD regularization for synthetic data, and sparser graph for real data.

Now we briefly outline several avenues for future work in this area. Although we observe empirically that the log-sum penalty significantly outperforms the lasso penalty, we have not proved it. This would be a task for future research. Theorem 1 shows that all three penalties considered yield the same asymptotic rate of convergence (see Remark 1) and how this rate is influenced by various chosen parameters. Other model selection approaches also need to be investigated. In \cite{Liu2010} a stability approach is proposed for graphical modeling with i.i.d.\ data using a random sampling approach to pick the smallest regularization parameter that ``simultaneously makes the graph sparse  and replicable under random sampling.'' In our case, in the frequency-domain we have $M$ models, each with $K$ approximately i.i.d.\ complex-valued Gaussian measurements ${\bm d}_x(\tilde{f}_{k,\ell})$ (see Sec.\ \ref{SM}-\ref{BPF1}). In order to apply the approach of \cite{Liu2010}, we would sample in the frequency-domain. Finally, sample complexity issues based on information-theoretic bounds need to be investigated. Based on the results of \cite{Wang2010} for i.i.d.\ data, \cite{Hannak2014} consider stationary Gaussian sequences, and using information-theoretic methods, \cite{Hannak2014} derive a lower bound on the error probability of any learning scheme for the underlying process CIG. This bound is then used to derive a minimum required sample-size which is necessary for any
algorithm regardless of its computational complexity, to reliably select the true underlying CIG. The model restrictions in \cite{Hannak2014} are more stringent than we consider here; e.g., instead of our assumption (A1), \cite{Hannak2014} needs $\sum_{\tau = -\infty}^\infty |\tau| \, | [{\bm R}_{xx}( \tau )]_{k \ell} | < \infty \mbox{ for every } 
		  k, \ell \in \bar{V}  \, .$

\section*{APPENDIX} 
\subsection{PROOF OF THEOREM 1} \label{append1}
Our proof relies on the method of \cite{Rothman2008} which deals with i.i.d.\ time series models and lasso penalty, and our prior results in \cite{Tugnait22c} dealing with sparse group lasso penalty and single-attribute time series. From now on we use the term ``with high probability'' (w.h.p.) to denote with probability greater than $1-1/(m p_n)^{\tau-2}$. First we recall Lemmas 2 and 3 from \cite{Tugnait22c}, replacing $p_n$ therein with $mp_n$.

We denote ${\bm S}_0(\tilde{f}_k)$ as ${\bm S}_{0k}$ in this section.  \\
{\it LEMMA 2}\cite[Lemma 3]{Tugnait22c}. Under Assumption (A1)-(A2), $\hat{\bm S}_k$ satisfies the tail bound
\begin{align*}  
   P \Big(  \max_{k,q,l} &\Big| [ \hat{\bm S}_k - {\bm S}_{0k} ]_{ql} \Big| 
	    > C_0 \sqrt{\frac{\ln(mp_n )}{K_n}} \Big)  \le \frac{1}{(mp_n) ^{\tau -2}} %\label{naeq58b}
\end{align*}
for $\tau > 2$, if the sample size  $n >  N_1$, where $C_0$ is defined in (\ref{naeq58}) and $N_1$ is defined in (\ref{neq15ab2}). $\quad \bullet$ 

Lemma 3 deals with a Taylor series expansion with integral remainder using Wirtinger calculus \cite{Schreier10}. \\
{\em LEMMA 3} \cite[Lemma 5]{Tugnait22c}. With $c(\bm{\Phi}_k, \bm{\Phi}_k^\ast) = \ln |\bm{\Phi}_k| + \ln |\bm{\Phi}_k^\ast|$ and  $\bm{\Phi}_k = \bm{\Phi}_{0k} + \bm{\Gamma}_k =\bm{\Phi}_k^H$, the Taylor series expansion of $c(\bm{\Phi}_k, \bm{\Phi}_k^\ast)$ in the integral remainder form is given by 
\begin{align}
   c& (\bm{\Phi}_k, \bm{\Phi}_k^\ast) =  c(\bm{\Phi}_{0k}, \bm{\Phi}_{0k}^\ast) 
	   +\mbox{tr}(\bm{\Phi}_{0k}^{-1} \bm{\Gamma}_k + \bm{\bm{\Phi}}_{0k}^{-\ast} \bm{\Gamma}_k^\ast ) \nonumber \\
		  & \;\; - \bm{g}^H(\bm{\Gamma}_k) \left( \int_0^1 (1-v) 
			  \bm{H}(\bm{\Phi}_{0k}, \bm{\Gamma}_k, v ) \, dv \right)  \bm{g}(\bm{\Gamma}_k)   \label{naeq100}
\end{align}
where $v$ is real, 
\begin{equation}
   \bm{g}(\bm{\Gamma}_k) = \left[ \begin{array}{c} \mbox{vec}(\bm{\Gamma}_k) \\
	       \mbox{vec}(\bm{\Gamma}_k^\ast)  \end{array} \right] , \;
  \bm{H}(\bm{\Phi}_{0k},\bm{\Gamma}_k, v ) 
	    = \left[ \begin{array}{cc} \bm{H}_{11k} &  \bm{0} \\
	       \bm{0} & \bm{H}_{22k}  \end{array} \right] 
\end{equation}
\begin{equation}
   \bm{H}_{11k} = (\bm{\Phi}_{0k} +v \bm{\Gamma}_k)^{-\ast} \otimes (\bm{\Phi}_{0k} +v \bm{\Gamma}_k)^{-1} \, ,
\end{equation}
and
\begin{equation}
  \bm{H}_{22k} = (\bm{\Phi}_{0k}+v \bm{\Gamma}_k)^{-1} \otimes (\bm{\Phi}_{0k}+v \bm{\Gamma}_k)^{-\ast}  \quad \bullet
\end{equation}

We now turn to the proof of Theorem 1. \\
{\em PROOF OF THEOREM 1}. Let $\bm{\Omega} = \bm{\Omega}_0 + \bm{\Delta}$ where
\begin{align}
   \bm{\Delta} & = \left[ \bm{\Gamma}_1 \; \bm{\Gamma}_2 \; \cdots \; \bm{\Gamma}_{M_n} \right] 
	  \in \mathbb{C}^{(mp_n) \times (mp_n M_n)} \, , \\
	 \bm{\Gamma}_k & = \bm{\Phi}_k - \bm{\Phi}_{0k} \in \mathbb{C}^{(mp_n) \times (mp_n )} , \; k \in [ M_n] ,
\end{align}
and $\bm{\Phi}_k$, $\bm{\Phi}_{0k}$ are both Hermitian positive-definite, implying $\bm{\Gamma}_k = \bm{\Gamma}_k^H$.
Let
\begin{equation}
  Q(\bm{\Omega}) := \bar{\cal L}(\bm{\Omega}) 
	           - \bar{\cal L}(\bm{\Omega}_0 ) \, .
\end{equation}
The estimate $\hat{\bm{\Omega}}_{\lambda}$, denoted by $\hat{\bm{\Omega}}$ hereafter suppressing dependence upon $\lambda$, minimizes $Q(\bm{\Omega})$, or equivalently, $\hat{\bm{\Delta}} = \hat{\bm{\Omega}} - \bm{\Omega}_0$ minimizes $G(\bm{\Delta}) := Q(\bm{\Omega}_0 + \bm{\Delta})$.
We will follow the proof of \cite[Theorem 1]{Tugnait22c}, which, in turn, follows the method of proof of \cite[Theorem 1]{Rothman2008} pertaining to real-valued i.i.d.\ time series. Consider the set
\begin{equation}  \label{naeq1001}
  \Theta_n(R) :=  \left\{ \bm{\Delta} \, :\, \bm{\Gamma}_k = \bm{\Gamma}_k^H , \;  k \in [M_n], \; \|\bm{\Delta} \|_F = R r_n \right\}
\end{equation}
where $R$ and $r_n$ are as in (\ref{neq15ab0}) and (\ref{neq15ab1}), respectively. Observe that 
\begin{equation}
     G(\hat{\bm{\Delta}}) = Q(\bm{\Omega}_0 + \hat{\bm{\Delta}}) \le G(\bm{0}) = 0 \, .
\end{equation} 
Therefore, if we can show that 
\begin{equation}
     \inf_{\bm{\Delta}}  \{ G(\bm{\Delta}) \, :\, \bm{\Delta} \in \Theta_n(R) \} \, > \, 0 \, ,
\end{equation}
a minimizer $\hat{\bm{\Delta}}$ must be inside the sphere defined by $\Theta_n(R)$, and hence
\begin{equation}
     \| \hat{\bm{\Delta}} \|_F \le   R r_n \, .
\end{equation}
When $G(\bm{\Delta})$ is convex (as with the lasso penalty), the minimizer is global, else it is local.

Using Lemma 3 we rewrite $G({\bm{\Delta}})$ as 
\begin{equation}
     G({\bm{\Delta}}) = \sum_{k=1}^{M_n} ( \frac{1}{2}A_{1k} + \frac{1}{2} A_{2k} + A_{3k}) + A_4 \, , \label{Gmain}
\end{equation}
where, noting that $\bm{\Phi}_{0k}^{-1} = \bm{S}_{0k}$, 
\begin{align} 
   &  A_{1k} =  \bm{g}^H(\bm{\Gamma}_k) \left( \int_0^1 (1-v) 
			  \bm{H}(\bm{\Phi}_{0k}, \bm{\Gamma}_k, v ) \, dv \right)  \bm{g}(\bm{\Gamma}_k)  \, ,  \label{naeq1100} \\
   &  A_{2k} =   \mbox{tr} \left( (\hat{\bm{S}}_k - \bm{S}_{0k} ) \bm{\Gamma}_k  + 
		  (\hat{\bm{S}}_k - \bm{S}_{0k} )^\ast \bm{\Gamma}_k^\ast \right) \, , \label{naeq1110} \\
   &  A_{3k} =  \alpha \sum_{i \ne j}^{mp_n} \Big( 
		  \rho_\lambda \big(  [ {\bm{\Phi}}_{0k}+\bm{\Gamma}_k ]_{ij} \big) -
			\rho_\lambda \big(  [ {\bm{\Phi}}_{0k} ]_{ij} \big) \Big) \, , \label{naeq1120}  \\
	&	A_{4} =  (1-\alpha) m \sqrt{M} \, \sum_{ q \ne \ell}^{p_n} \Big( 
		    \rho_\lambda \big( \| {\bm \Omega}_0^{(q \ell M_n)} + \bm{\Delta}^{(q \ell M_n)} \|_F \big) \nonumber \\
		& \quad\quad\quad		- \rho_\lambda \big( \| {\bm \Omega}_0^{(q \ell M_n)} \|_F \big) \Big) \, , \label{naeq1122} \\
  &  {\bm \Omega}_0^{(q \ell M_n)} :=  [ {\bm{\Phi}}_1^{(q \ell)} , \; {\bm{\Phi}}_2^{(q \ell)} ,
   \; \cdots ,\; {\bm{\Phi}}_{M_n}^{(q \ell)}]
			  \in \mathbb{C}^{m \times (mM_n)} \, ,  \label{naeq1100a} \\
  &   \bm{\Delta}^{(q \ell M_n)} :=   [ {\bm{\Gamma}}_1^{(q \ell)} , \; {\bm{\Gamma}}_2^{(q \ell)} ,
   \; \cdots ,\; {\bm{\Gamma}}_{M_n}^{(q \ell)}]
			  \in \mathbb{C}^{m \times (mM_n)} \, . \label{naeq1110b} 
\end{align}
Also define
\begin{equation}
     A_\ell=0.5 \sum_{k=1}^{M_n} A_{\ell k}, \; \ell =1,2, \;\; A_3 = \sum_{k=1}^{M_n} A_{3 k} \, ,
\end{equation}
and
\begin{equation}
  d_{1n} := \sqrt{\frac{\ln(mp_n ) }{ K_n} }, \;\;
	d_{2n} : = d_{1n} \sqrt{ mp_n+m^2s_{n0} } \, .
\end{equation}
The bounds on $A_{1k}$'s and $A_1$ follows exactly as in \cite[Theorem 1]{Tugnait22c}, with the final result (see equations \cite[(B.39)-(B.44)]{Tugnait22c})
\begin{equation}
    A_1 \ge  \frac{\| \bm{\Delta} \|_F^2 }
		  {2 \left(\beta_{\min}^{-1} + R r_n \right)^{2}} \, .  \label{A1main}
\end{equation}

Turning to $A_{2k}$'s and $A_2$, as in \cite[Theorem 1]{Tugnait22c} (after accounting for the fact that here we have $\hat{\bm{S}}_k \in \mathbb{C}^{(mp_n) \times (mp_n)}$ whereas in \cite{Tugnait22c}, $\hat{\bm{S}}_k \in \mathbb{C}^{p_n \times p_n}$, and here we have group penalty on groups of size $m^2 M_n$ elements whereas in \cite{Tugnait22c}, group size is $M_n$), with probability $> \, 1-1/(mp_n)^{\tau-2}$, we have the bound \cite[(B.51)]{Tugnait22c}
\begin{align}
  |A_2| & \le  
	  C_0 \, \sum_{k=1}^{M_n} \Big( d_{1n}  \| \bm{\Gamma}_k^- \|_1  
	    +  d_{2n}  \| \bm{\Gamma}_k^+ \|_F \Big)  \label{naeq1200}
\end{align}
a well as the bound \cite[(B.55)]{Tugnait22c}
\begin{align}
  |A_2| & \le   \sqrt{m^2 M_n} \, C_0 \,  d_{1n} 
		   \big( \| \tilde{\bm{\Delta}}^-\|_1 + \|\tilde{\bm{\Delta}}^+\|_1 \big) \label{naeq1201}
\end{align}
where $\tilde{\bm{\Delta}} \in \mathbb{R}^{p_n \times p_n}$ has its $(i,j)$th element $\tilde{{\Delta}}_{ij} = \| {\bm \Delta}^{(ij M_n)} \|_F$.

For the rest of the proof we have two slightly different approaches, one for lasso and log-sum and the other for SCAD penalty. The following applies to lasso and log-sum penalties. \\
{\it For Lasso and Log-Sum Penalties}: 
We now bound $A_{3k}$. Let ${\cal E}_0^c$ denote the complement of ${\cal E}_0$, given by ${\cal E}_0^c = \{ \{i,j\} ~:~ ({\bm S}^{-1}_{0}(f))^{(ij)} \equiv {\bm 0}, ~i\ne j,  ~ i,j \in [p_n], ~ f \in [0,0.5] \}$. Similarly, let $\bar{\cal E}_0^c$ denote the complement of $\bar{\cal E}_0$, given by $\bar{\cal E}_0^c = \{ \{i,j\} ~:~ [{\bm S}^{-1}_{0}(f)]_{ij} \equiv 0, ~i\ne j, ~ i,j \in [mp_n], ~ f \in [0,0.5] \}$. For an index set ${\bm B}$ and a matrix ${\bm C} \in \mathbb{C}^{q \times q}$, we write ${\bm C}_{\bm B}$ to denote a matrix in $\mathbb{C}^{q \times q}$ such that $[{\bm C}_{\bm B}]_{ij} = C_{ij}$ if $(i,j) \in {\bm B}$, and $[{\bm C}_{\bm B}]_{ij}=0$ if $(i,j) \not\in {\bm B}$. Then $\bm{\Gamma}_k^- = \bm{\Gamma}_{k{\cal E}_0}^- + \bm{\Gamma}_{k{\cal E}_0^c}^-$, and $\| \bm{\Gamma}_k^- \|_1 = \| \bm{\Gamma}_{k{\cal E}_0}^- \|_1 + \| \bm{\Gamma}_{k{\cal E}_0^c}^- \|_1 $. We have
\begin{align}
     A_{3k} & =  \alpha \sum_{(i,j) \in \bar{\cal E}_0} \Big( 
		  \rho_\lambda \big(  [ {\bm{\Phi}}_{0k}+\bm{\Gamma}_k ]_{ij} \big) -
			\rho_\lambda \big(  [ {\bm{\Phi}}_{0k} ]_{ij} \big) \Big)  \nonumber \\
		&	\quad\quad\quad + \alpha \sum_{(i,j) \in \bar{\cal E}_0^c}  
			\rho_\lambda \big(  [ \bm{\Gamma}_k ]_{ij} \big)  \nonumber \\
		& =  \alpha \sum_{(i,j) \in \bar{\cal E}_0}  \rho_\lambda^\prime \big(  [ \tilde{\bm{\Phi}}_{k} ]_{ij} \big) 
		  \Big( |[ {\bm{\Phi}}_{0k}+\bm{\Gamma}_k ]_{ij}|- |[ {\bm{\Phi}}_{0k} ]_{ij}| \Big)  \nonumber \\
		&	\quad\quad\quad + \alpha \sum_{(i,j) \in \bar{\cal E}_0^c}  
			\rho_\lambda \big(  [ \bm{\Gamma}_k ]_{ij} \big)   \label{expan}
\end{align}
where we used the mean value theorem to infer $\rho_\lambda \big(  [ {\bm{\Phi}}_{0k}+\bm{\Gamma}_k ]_{ij} \big) = \rho_\lambda \big(  [ {\bm{\Phi}}_{0k} ]_{ij} \big) + \rho_\lambda^\prime \big(  [ \tilde{\bm{\Phi}}_{k} ]_{ij} \big) \big( |[ {\bm{\Phi}}_{0k}+\bm{\Gamma}_k ]_{ij}|- |[ {\bm{\Phi}}_{0k} ]_{ij}| \big)$ for some $|[ \tilde{\bm{\Phi}}_{k} ]_{ij}| = |[ {\bm{\Phi}}_{0k} ]_{ij}| + \gamma \Big( |[ {\bm{\Phi}}_{0k}+\bm{\Gamma}_k ]_{ij}|- |[ {\bm{\Phi}}_{0k} ]_{ij}| \Big)$ and $\gamma \in [0,1]$. Using the triangle inequality,  properties (vii) and (viii) of the penalty functions, and $C_\lambda = \lambda/2$, we have
\begin{align}
     A_{3k} & \ge  -  \alpha \sum_{(i,j) \in \bar{\cal E}_0}  \rho_\lambda^\prime \big(  [ \tilde{\bm{\Phi}}_{k} ]_{ij} \big) 
		   \, |[ \bm{\Gamma}_k ]_{ij}|  \nonumber \\
			& \quad \; + \alpha \sum_{(i,j) \in \bar{\cal E}_0^c}  
			C_\lambda \, |[ \bm{\Gamma}_k ]_{ij}|  \;\; \mbox{for } \; |[ \bm{\Gamma}_k ]_{ij}| \le \delta_\lambda \label{lowerbnd} \\
%\end{align}
%\vspace*{-0.15in}
%\begin{align}
		& \ge  -  \alpha \lambda_n \sum_{(i,j) \in \bar{\cal E}_0}   |[ \bm{\Gamma}_k ]_{ij}|  
		+ \frac{\alpha \lambda_n}{2} \sum_{(i,j) \in \bar{\cal E}_0^c}   |[ \bm{\Gamma}_k ]_{ij}| \nonumber \\
			& =  \alpha \lambda_n ( \frac{1}{2} \|\bm{\Gamma}_{k \bar{\cal E}_0^c}^- \|_1 
			            - \| \bm{\Gamma}_{k \bar{\cal E}_0}^- \|_1 )  \, ,
\end{align}
leading to ($A_3 = \sum_{k=1}^{M_n} A_{3k}$)
\begin{align}
     A_{3} 
			& \ge  \alpha \lambda_n \sum_{k=1}^{M_n} 
			( \frac{1}{2} \|\bm{\Gamma}_{k \bar{\cal E}_0^c}^- \|_1 - \| \bm{\Gamma}_{k \bar{\cal E}_0}^- \|_1 )  \, .
			  \label{naeq9202}
\end{align}
Similarly, by (\ref{naeq1122}), we have
\begin{align}
     A_{4} 
			& \ge  (1-\alpha) m \sqrt{M_n} \lambda_n \Big( \frac{1}{2} \sum_{ (q, \ell) \in {\cal E}_0^c }
			 \|  \bm{\Delta}^{(q \ell M_n)} \|_F  \nonumber \\
			& \quad\quad - \sum_{ (q, \ell) \in {\cal E}_0 }
			 \|  \bm{\Delta}^{(q \ell M_n)} \|_F \Big)  \, .  \label{naeq9203}
\end{align}

Now $\| \bm{\Gamma}_{k \bar{\cal E}_0}^- \|_1 \le \sqrt{m^2 s_{n0}} \, \| \bm{\Gamma}_{k \bar{\cal E}_0}^- \|_F \le \sqrt{m^2 s_{n0}} \, \| \bm{\Gamma}_{k} \|_F$, by the Cauchy-Schwarz inequality, hence
\begin{equation}
  \sum_{k=1}^{M_n} \| \bm{\Gamma}_{k \bar{\cal E}_0}^- \|_1 \le \sqrt{M_n m^2 s_{n0}} \| {\bm \Delta} \|_F \, .
	  \label{naeq9210}
\end{equation}
Set $\| \bm{\Gamma}_k^- \|_1 = \| \bm{\Gamma}_{k \bar{\cal E}_0}^- \|_1 + \| \bm{\Gamma}_{k \bar{\cal E}_0^c}^- \|_1 $ in $A_2$ of (\ref{naeq1200}) to deduce that w.h.p.\
\begin{align}
  \alpha & A_2 + A_3  \ge  -\alpha |A_2| + A_3 \nonumber \\
			& \ge \alpha(0.5 \lambda_n - C_0 d_{1n} ) 
	           \sum_{k=1}^{M_n} \| \bm{\Gamma}_{k \bar{\cal E}_0^c}^- \|_1  \nonumber \\
			& \quad\quad - \alpha ( C_0 d_{1n} + \lambda_n) \sum_{k=1}^{M_n} \| \bm{\Gamma}_{k \bar{\cal E}_0}^- \|_1
			   - \alpha  C_0 d_{2n} \sum_{k=1}^{M_n} \| \bm{\Gamma}_{k}^+ \|_F \nonumber \\
%\end{align}
%\vspace*{-0.15in}
%\begin{align} 
			& \ge - \alpha \Big( ( C_0 d_{1n} + \lambda_n) \sqrt{m^2 s_{n0}} 
			   +  C_0 d_{2n} \Big) \sqrt{M_n } \| {\bm \Delta} \|_F  \nonumber \\
			& \ge - \alpha \Big( m \sqrt{s_{n0}} \, \lambda_n  
			   +  2 C_0 d_{2n} \Big) \sqrt{M_n } \, \| {\bm \Delta} \|_F \label{naeq9212}
\end{align} 
where we have used the fact that $0.5 \lambda_n \ge    C_0 d_{1n} = \lambda_{n\ell}/2$ (see (\ref{neq15ab5})), (\ref{naeq9210}),  $\sum_{k=1}^{M_n} \| \bm{\Gamma}_{k}^+ \|_F \le \sqrt{M_n } \, \| {\bm \Delta} \|_F$ (by the Cauchy-Schwarz inequality), and the bound $\sqrt{s_{n0}} \, m d_{1n} \le d_{2n}$. Now use $A_2$ of (\ref{naeq1201}) to deduce that w.h.p.\
\begin{align}
  & (1-\alpha)  A_2 + A_4  \ge  -(1-\alpha) |A_2| + A_4 \nonumber \\
			& \ge (1-\alpha) m \sqrt{M_n}  (0.5 \lambda_n - C_0 d_{1n} ) \sum_{ (q, \ell) \in {\cal E}_0^c }
			 \|  \bm{\Delta}^{(q \ell M_n)} \|_F  \nonumber \\
			& \quad - (1-\alpha) m \sqrt{M_n} \Big( ( C_0 d_{1n} + \lambda_n) \sum_{ (q, \ell) \in {\cal E}_0 }
			 \|  \bm{\Delta}^{(q \ell M_n)} \|_F  \nonumber \\
			& \quad\quad +  C_0 d_{1n} \sum_{ q = \ell=1}^{p_n} \|  \bm{\Delta}^{(q \ell M_n)} \|_F \Big) \nonumber \\
			& \ge - (1-\alpha) \sqrt{M_n } \, \| {\bm \Delta} \|_F  \, 
			\Big( \sqrt{m^2 s_{n0}} \, \lambda_n   \nonumber \\
			& \quad\quad \quad +  C_0 d_{1n} m \big(  \sqrt{s_{n0}} +  \sqrt{p_n} \big) \Big) \nonumber \\
			& \ge - (1-\alpha) \Big( m \sqrt{s_{n0}} \, \lambda_n +  C_0 (1+m)d_{2n} \Big) 
			   \sqrt{M_n } \, \| {\bm \Delta} \|_F \label{naeq9215}
\end{align}
where we have used the facts that $0.5 \lambda_n \ge    C_0 d_{1n} = \lambda_{n\ell}/2$, $\sum_{ (q, \ell) \in {\cal E}_0 } \|  \bm{\Delta}^{(q \ell M_n)} \|_F \le \sqrt{s_{n0}} \| {\bm{\Delta}} \|_F$ and $\sum_{ q = \ell=1}^{p_n}\|  \bm{\Delta}^{(q \ell M_n)} \|_F \le \sqrt{p_n} \| {\bm{\Delta}} \|_F$ by the Cauchy-Schwarz inequality, and the bounds $\sqrt{s_{n0}} \, m d_{1n} \le d_{2n}$ and $\sqrt{ p_n} \,m  d_{1n} \le m d_{2n}$. 

From (\ref{naeq9212}) and (\ref{naeq9215}), after some simplifications, we have
\begin{align}
  & A_2 +  A_3 + A_4  \ge -  \Big( m \sqrt{s_{n0}} \, \lambda_n +  C_0 (1+m)d_{2n} \Big) \nonumber \\
	 & \quad\quad \times \sqrt{M_n } \, \| {\bm{\Delta}} \|_F
	   \label{Armain0}
\end{align}
where we used the bound $2 d_{2n} \le (1+m)d_{2n}$. By (\ref{neq15ab1}), (\ref{neq15ab6}) and (\ref{neq15ab7}), $\lambda_n$ is chosen to satisfy 
\begin{align}
  \lambda_n & \le \lambda_{nu1} = \frac{C_0 (1+m)}{m \sqrt{s_{n0} M_n}}\, r_n \, .
\end{align}
Noting that $r_n = \sqrt{M_n} \, d_{2n}$, we have
\begin{align}
  & A_2 +  A_3 + A_4  \ge -  2 C_0 (1+m) r_n  \| {\bm{\Delta}} \|_F \, .
	   \label{Armain}
\end{align}
Using (\ref{Gmain}), (\ref{A1main}) and (\ref{Armain}), and $\| {\bm \Delta} \|_F = R r_n$, we have w.h.p.\
\begin{align}
     G({\bm{\Delta}}) \ge &  \; \| \bm{\Delta} \|_F^2 \left[ \frac{1}{2 (\beta_{\min}^{-1} + R r_n )^2}  
		     - \frac{2C_0(1+m)}{R} \right]  \, .\label{naeq1370} 
\end{align}
For the given choice of $N_2$, $R r_n \le R r_{N_2} \le 0.1/\beta_{\min}$ for $n \ge N_2$. Also, $2C_0(1+m)/R = \beta_{\min}^2/4$ by (\ref{neq15ab0}). Then for $n \ge N_2$,
\begin{align*}
    \frac{1}{2 (\beta_{\min}^{-1} + R r_n )^2}- \frac{2C_0(1+m)}{R}
		\ge \beta_{\min}^2 \left( \frac{1}{2.42} - \frac{1}{4} \right) > 0 \, ,
\end{align*}
implying $G({\bm{\Delta}})  > 0$. This proves (\ref{neq15}).  The choice of $N_3$ for log-sum penalty ensures that $| [{\bm \Gamma}_k]_{ij} | \le \delta_\lambda = \epsilon$  needed in (\ref{lowerbnd}) is satisfied w.h.p.: if $R r_n \le  \epsilon$, then  $| [{\bm \Gamma}_k]_{ij} | \le  \|{\bm \Delta}\|_F \le R r_n \le \epsilon$. 

The following applies to the SCAD penalty. \\
{\it For SCAD Penalty}: Here we address (\ref{expan}) differently. Using triangle inequality, we have 
\begin{align}
  |[ \tilde{\bm{\Phi}}_{k} ]_{ij}| & \ge |[ {\bm{\Phi}}_{0k} ]_{ij}| 
 + \gamma \Big( |[ {\bm{\Phi}}_{0k} ]_{ij}| - |[\bm{\Gamma}_k ]_{ij}|- |[ {\bm{\Phi}}_{0k} ]_{ij}| \Big) \nonumber \\
	  & \ge |[ {\bm{\Phi}}_{0k} ]_{ij}| - |[\bm{\Gamma}_k ]_{ij}| \, . \label{expan2}
\end{align}
Since $| [\bm{\Gamma}_k ]_{ij} | \le \| \bm{\Delta} \|_F \le R r_n$, the choice $\lambda_n = \lambda_{nu2}$ implies that $\lambda_n \ge Rr_n$, satisfying $| [ \bm{\Gamma}_k ]_{ij} | \le  \lambda_n$. Therefore, $|[ \tilde{\bm{\Phi}}_{k} ]_{ij} \ge |[ {\bm{\Phi}}_{0k} ]_{ij}| - \lambda_n$.  For $n \ge N_4$, $\rho_\lambda^\prime (|[ \tilde{\bm{\Phi}}_{k} ]_{ij}|)=0$ (see (\ref{neq15ab4a}) if $\{i,j\} \in \bar{\cal E}_0$, i.e, $[ {\bm{\Phi}}_{0k} ]_{ij} \ne 0$, since in this case $|[ \tilde{\bm{\Phi}}_{k} ]_{ij}| \ge (a+1) \lambda_n  - \lambda_n = a \lambda_n$. Therefore, for $n \ge N_4$, 
\begin{align}
     A_{3k} & =  \alpha \sum_{(i,j) \in \bar{\cal E}_0^c}  
			\rho_\lambda \big(  [ \bm{\Gamma}_k ]_{ij} \big) \nonumber \\
		& \ge \alpha \sum_{(i,j) \in \bar{\cal E}_0^c}  
			C_\lambda \, |[ \bm{\Gamma}_k ]_{ij}|  \;\; \mbox{for } \; |[ \bm{\Gamma}_k ]_{ij}| \le \delta_\lambda 
\nonumber \\
			& =  \alpha (\lambda_n /2) \|\bm{\Gamma}_{k \bar{\cal E}_0^c}^- \|_1  \label{eqnn22}
\end{align}
leading to ($A_3 = \sum_{k=1}^{M_n} A_{3k}$)
\begin{align}
     A_{3} 
			& \ge  \alpha \, (\lambda_n /2) \sum_{k=1}^{M_n} 
			 \|\bm{\Gamma}_{k \bar{\cal E}_0^c}^- \|_1   \, .
			  \label{naeq92022}
\end{align} 
Mimicking the steps for bounding $A_3$ above and under the same conditions, we have
\begin{align}
     A_{4} 
			& \ge  (1-\alpha) m \sqrt{M_n}\,  (\lambda_n /2)  \sum_{ (q, \ell) \in {\cal E}_0^c }
			 \|  \bm{\Delta}^{(q \ell M_n)} \|_F    \, .  \label{naeq920322}
\end{align}
Thus w.h.p.\
\begin{align}
  \alpha & A_2 + A_3  \ge  -\alpha |A_2| + A_3 \nonumber \\
			& \ge \alpha(0.5 \lambda_n - C_0 d_{1n} ) 
	           \sum_{k=1}^{M_n} \| \bm{\Gamma}_{k \bar{\cal E}_0^c}^- \|_1  \nonumber \\
			& \quad\quad - \alpha  C_0 d_{1n}  \sum_{k=1}^{M_n} \| \bm{\Gamma}_{k \bar{\cal E}_0}^- \|_1
			   - \alpha  C_0 d_{2n} \sum_{k=1}^{M_n} \| \bm{\Gamma}_{k}^+ \|_F \nonumber \\
%\end{align}
%\vspace*{-0.15in}
%\begin{align} 
			& \ge - \alpha \Big(  C_0 d_{1n}  \sqrt{m^2 s_{n0}} 
			   +  C_0 d_{2n} \Big) \sqrt{M_n } \| {\bm \Delta} \|_F  \nonumber \\
			& \ge - \alpha 2 C_0 d_{2n}  \sqrt{M_n } \, \| {\bm \Delta} \|_F \label{naeq92122}
\end{align} 
where we have used (\ref{naeq9210}),  $\sum_{k=1}^{M_n} \| \bm{\Gamma}_{k}^+ \|_F \le \sqrt{M_n } \, \| {\bm \Delta} \|_F$ (by the Cauchy-Schwarz inequality), the bound $\sqrt{s_{n0}} \, m d_{1n} \le d_{2n}$, and  the fact since $\lambda_n = \max \left( R r_n , \lambda_{n u1} \right)$ in Theorem 1, $0.5 \lambda_n - C_0 d_{1n} \ge    0$ and therefore, the term involving $0.5 \lambda_n - C_0 d_{1n}$ above can be neglected. By very similar arguments we also have
\begin{align}
  & (1-\alpha)  A_2 + A_4  \ge  -(1-\alpha) |A_2| + A_4 \nonumber \\
			& \ge (1-\alpha) m \sqrt{M_n}  (0.5 \lambda_n - C_0 d_{1n} ) \sum_{ (q, \ell) \in {\cal E}_0^c }
			 \|  \bm{\Delta}^{(q \ell M_n)} \|_F  \nonumber \\
			& \quad - (1-\alpha) m \sqrt{M_n} \Big(  C_0 d_{1n}  \sum_{ (q, \ell) \in {\cal E}_0 }
			 \|  \bm{\Delta}^{(q \ell M_n)} \|_F  \nonumber \\
			& \quad\quad +  C_0 d_{1n} \sum_{ q = \ell=1}^{p_n} \|  \bm{\Delta}^{(q \ell M_n)} \|_F \Big) \nonumber \\
			& \ge - (1-\alpha) \sqrt{M_n } \, \| {\bm \Delta} \|_F  \, 
			\Big( C_0 d_{1n} m ( \sqrt{s_{n0}} +  \sqrt{p_n} \,)  \Big) \nonumber \\
			& \ge - (1-\alpha)  C_0 (1+m)d_{2n}  
			   \sqrt{M_n } \, \| {\bm \Delta} \|_F \label{naeq92152}
\end{align}
where we have used the facts that $0.5 \lambda_n \ge    C_0 d_{1n} = \lambda_{n\ell}/2$, $\sum_{ (q, \ell) \in {\cal E}_0 } \|  \bm{\Delta}^{(q \ell M_n)} \|_F \le \sqrt{s_{n0}} \| {\bm{\Delta}} \|_F$ and $\sum_{ q = \ell=1}^{p_n}\|  \bm{\Delta}^{(q \ell M_n)} \|_F \le \sqrt{p_n} \| {\bm{\Delta}} \|_F$ by the Cauchy-Schwarz inequality, and the bounds $\sqrt{s_{n0}} \, m d_{1n} \le d_{2n}$ and $\sqrt{ p_n} \,m  d_{1n} \le m d_{2n}$. From (\ref{naeq92122}) and (\ref{naeq92152}) we have
\begin{align}
  A_2+ & A_3+A_4 \ge  - C_0 (1+m)d_{2n}  
			   \sqrt{M_n } \, \| {\bm \Delta} \|_F \nonumber \\
		 & \ge -  C_0  (1+ m)  r_n \| \bm{\Delta} \|_F \label{Armain2}
\end{align}
where we used $r_n = \sqrt{M_n} \, d_{2n}$ and the bound $2 d_{2n} \le (1+m)d_{2n}$.
Mimicking (\ref{naeq1370}), we have with probability $> 1- 1/(mp_n)^{\tau-2}$, we have
\begin{align}
     G({\bm{\Delta}}) \ge &  \; \| \bm{\Delta} \|_F^2 \left[ \frac{1}{2 (\beta_{\min}^{-1} + R r_n )^2}  
		     - \frac{C_0(1+m)}{R} \right]  \nonumber \\
				\ge& \beta_{\min}^2 \left( \frac{1}{2.42} - \frac{1}{8} \right) > 0 \, , \label{naeq13702} 
\end{align}
implying $G({\bm{\Delta}})  > 0$. This proves (\ref{neq15}). For the SCAD penalty, we  need $| [{\bm \Gamma}_k]_{ij} |  \le \delta_\lambda = \lambda_n$ in (\ref{eqnn22}). Since $| [{\bm \Gamma}_k]_{ij} | \le \| \bm{\Delta} \|_F \le R r_n$, the choice $\lambda_n = \lambda_{nu2}$ implies that $\lambda_n \ge Rr_n$, satisfying $| [{\bm \Gamma}_k]_{ij} | \le  \lambda_n$. This completes the proof.   $\quad \blacksquare$

\subsection{PROOFS OF LEMMA 1 and THEOREMS 2 and 3} \label{append2}

{\em PROOF OF LEMMA 1}. Consider ${\cal L}({\bm \Omega})- \frac{\mu}{2} \| {\bm \Omega} \|_F^2$ for some $\mu \ge 0$. By \cite[Lemma 4]{Tugnait22c}, using Wirtinger calculus, the Hessian of ${\cal L}({\bm \Omega})$ w.r.t.\ 
\begin{align*}
    {\bm y} = & \big[ (\mbox{vec}({\bm \Phi}_1))^\top , \; (\mbox{vec}({\bm \Phi}_1^\ast))^\top , \; \cdots , \\
		  & \quad\quad (\mbox{vec}({\bm \Phi}_{M_n}))^\top , \; (\mbox{vec}({\bm \Phi}_{M_n}^\ast))^\top \big]^\top 
			      \in \mathbb{C}^{2 m^2p_n^2 M_n}
\end{align*}
is given by
\begin{align}
     \nabla^2 {\cal L}({\bm \Omega}) = &   \mbox{block-diag}\Big\{
		  {\bm \Phi}_1^{-\ast} \otimes {\bm \Phi}_1^{-1}, \; {\bm \Phi}_1^{-1} \otimes {\bm \Phi}_1^{-\ast} , \; \cdots , \nonumber \\
		& \quad\quad  {\bm \Phi}_{M_n}^{-\ast} \otimes {\bm \Phi}_{M_n}^{-1}, \; {\bm \Phi}_{M_n}^{-1} \otimes {\bm \Phi}_{M_n}^{-\ast} \Big\} 
			 \label{naeq8370} 
\end{align}
with 
\begin{align}
     \phi_{\min}(\nabla^2 {\cal L}({\bm \Omega})) = &  \min_k \phi_{\min}^2( {\bm \Phi}_k^{-1})
		  =  \frac{1}{\max_k \phi_{\max}^2( {\bm \Phi}_k)} \nonumber \\
		 = & \frac{1}{\max_k \| {\bm \Phi}_k \|^2 }  \ge \beta_{\min}^2 \, .
			 \label{naeq8372} 
\end{align}
Since we have $\| {\bm \Omega} \|_F^2 = \frac{1}{2} {\bm y}^H {\bm y}$, the Hessian of $\| {\bm \Omega} \|_F^2$ w.r.t.\ ${\bm y} $
is given by
\begin{align}
     \nabla^2 \| {\bm \Omega} \|_F^2 = &  {\bm I}_{2 m^2p_n^2 M_n} \, .
			 \label{naeq8374} 
\end{align}
It follows from (\ref{naeq8372}) and (\ref{naeq8374}) that ${\cal L}({\bm \Omega})- \frac{\mu}{2} \| {\bm \Omega} \|_F^2$  is positive semi-definite, hence convex, if  
\begin{align}
     \| {\bm \Phi}_k \| & \le \sqrt{\frac{2}{\mu}} \, \;\; \forall k \in [M_n] \, .
			 \label{naeq8375} 
\end{align}
By property (v) of the penalty functions, $g(u):=\rho_\lambda(u) +\frac{\mu}{2} u^2$ is convex, for some $\mu \ge 0$, and by property (ii), it is non-decreasing on $\mathbb{R}_+$. Therefore, by the composition rules \cite[Sec.\ 3.2.4]{Boyd2004}, $g(| [ {\bm{\Phi}}_k ]_{ij}|)$ and $g(\| {\bm \Omega}^{(q \ell M_n)} \|_F)$ are convex. Hence,
\begin{align}
     &  P_e({\bm \Omega}) + \frac{\mu_e}{2} \| {\bm \Omega} \|_F^2  
		 =   \sum_{k=1}^{M_n} \; \sum_{i \ne j}^{mp} 
		 \Big( \rho_\lambda(\big| [ {\bm{\Phi}}_k ]_{ij} \big|) +\frac{\mu_e}{2} \big| [ {\bm{\Phi}}_k ]_{ij} \big|^2 \Big)	 
		  \label{naeq8378}
\end{align}
is convex for $\mu_e = \mu \ge 0$, and similarly,
\begin{align}
  &   P_g({\bm \Omega}) + \frac{\mu_g}{2} \| {\bm \Omega} \|_F^2  
		 =  m \sqrt{M_n} \sum_{k=1}^{M_n} \; \sum_{ q \ne \ell}^p \; 
		   \Big( \rho_\lambda ( \| {\bm \Omega}^{(q \ell M_n)} \|_F )  \nonumber \\
		& \quad\quad	+\frac{\mu_g}{2 m \sqrt{M_n}} \| {\bm \Omega}^{(q \ell M_n)} \|_F^2 \Big)	
				  \label{naeq8379}
\end{align}
is convex for $\mu_g = m \sqrt{M_n} \, \mu$, where $\mu$ is the value that renders $\rho_\lambda(u) +\frac{\mu}{2} u^2$ convex. Now express $\bar{\cal L}({\bm \Omega})$ as
\begin{align}
   \bar{\cal L}({\bm \Omega}) = & \alpha \bar{\cal L}_e({\bm \Omega})  
	 + (1-\alpha) \bar{\cal L}_g({\bm \Omega}) \, , \label{naeq8381} \\
	\bar{\cal L}_e({\bm \Omega}) = & 	{\cal L}({\bm \Omega}) - \frac{\mu}{2} \| {\bm \Omega} \|_F^2
	  + P_e({\bm \Omega}) + \frac{\mu}{2} \| {\bm \Omega} \|_F^2  \, , \label{naeq8381a} \\
		\bar{\cal L}_g({\bm \Omega}) = & {\cal L}({\bm \Omega}) - \frac{\mu}{2} \| {\bm \Omega} \|_F^2
	  + P_g({\bm \Omega}) + \frac{\mu}{2} \| {\bm \Omega} \|_F^2   \, .
				  \label{naeq8381b}
\end{align}
By (\ref{naeq8375}), (\ref{naeq8378}) and (\ref{naeq8381a}), $\bar{\cal L}_e({\bm \Omega})$ is convex function of ${\bm \Omega}$ if
\begin{align}
     \| {\bm \Phi}_k \| & \le \sqrt{\frac{2}{\mu}} \, \;\; \forall k \in [M_n] \, ,
			 \label{naeq8385} 
\end{align}
and by (\ref{naeq8375}), (\ref{naeq8379}) and (\ref{naeq8381b}), $\bar{\cal L}_g({\bm \Omega})$ is convex in ${\bm \Omega}$ if
\begin{align}
     \| {\bm \Phi}_k \| & \le \sqrt{\frac{2}{\mu_g}} = \sqrt{\frac{2}{m \mu \sqrt{M_n}}} \, \;\; \forall k \in [M_n] \, .
			 \label{naeq8387} 
\end{align}
Thus, for $\bar{\cal L}({\bm \Omega})$ to be strictly convex, using the (minimum) values of $\mu$ to make $\rho_\lambda(u) +\frac{\mu}{2} u^2$ convex, we require
\begin{align}  
		   \| \bm{\Phi}_{k} \| < & \sqrt{\frac{2}{m \mu \sqrt{M_n}}} \, \;\; \forall k \in [M_n] \nonumber \\
		 = & \left\{ \begin{array}{ll}
		   \infty & : \;\; \mbox{Lasso} \\
			  \sqrt{\frac{2(a-1)}{m \sqrt{M_n}}} & : \;\; \mbox{SCAD} \\
				\sqrt{ \frac{ 2 \epsilon}{m \sqrt{M_n} \lambda_n} } & : \;\; \mbox{log-sum}, \end{array} \right. \label{naeq8390} 
\end{align}
The choice $\| \bm{\Phi}_{k} \| <  \sqrt{\frac{2}{m \mu \sqrt{M_n}}}$ makes ${\cal L}({\bm \Omega})- \frac{\mu}{2} \| {\bm \Omega} \|_F^2$ positive definite, hence strictly convex. We take $\| \bm{\Phi}_{k} \| = 0.99 \,  \sqrt{\frac{2}{m \mu \sqrt{M_n}}}$, completing the proof.  $\quad \blacksquare$

{\em PROOF OF THEOREM 2}. If $1/\beta_{\min} \le  0.99 \,  \sqrt{\frac{2}{m \mu \sqrt{M_n}}}$, then $\bm{\Phi}_{0k}  \in {\cal B}_k$ since $\| \bm{\Phi}_{0k} \| = 1/ \phi_{\min}(\bm{S}_{0k} ) \le 1/\beta_{\min}$ by assumption (A4). To establish that  $\hat{\bm{\Phi}}_{k}  \in {\cal B}_k$, consider (${\bm \Delta}$ is as in the proof of Theorem 1)
\begin{align}  
		   \| \hat{\bm \Phi}_{k} \| \le & \| \hat{\bm \Phi}_{k} - {\bm \Phi}_{0k} \|
			  + \| {\bm \Phi}_{0k} \|  \nonumber \\
		 \le & \| {\bm \Delta} \| + 1/\beta_{\min} \le \| {\bm \Delta} \|_F + 1/\beta_{\min} \nonumber \\
		 \le & R r_n + 1/\beta_{\min} \, .
		\label{naeq8394} 
\end{align}
Therefore, $\hat{\bm{\Phi}}_{k}  \in {\cal B}_k$. Thus, both $\hat{\bm{\Phi}}_{k}$ and $\bm{\Phi}_{0k}$, hence $\hat{\bm{\Omega}}_\lambda$ and ${\bm{\Omega}}_0$, respectively, are feasible. The desired result then follows from Theorem 1 and (local) strict convexity of $\bar{\cal L}({\bm \Omega})$ over $\cap_{k=1}^{M_n} \,{\cal B}_k$ implied by Lemma 1. $\quad \blacksquare$

{\em PROOF OF THEOREM 3}. We have $\| \hat{\bm \Omega}^{(q \ell M_n)} - {\bm \Omega}_0^{(q \ell M_n)} \|_F \le \| \hat{\bm \Omega} - {\bm \Omega}_0 \|_F \le \bar{\sigma}_n$ w.h.p. For the edge $\{ q,\ell \} \in {\cal E}_0$, we have
\begin{align}  
		   \| \hat{\bm \Omega}^{(q \ell M_n)} \|_F = & 
			    \| {\bm \Omega}_0^{(q \ell M_n)} + \hat{\bm \Omega}^{(q \ell M_n)} - {\bm \Omega}_0^{(q \ell M_n)} \|_F  \nonumber \\
		 \ge & \| {\bm \Omega}_0^{(q \ell M_n)} \|_F - \| \hat{\bm \Omega}^{(q \ell M_n)} - {\bm \Omega}_0^{(q \ell M_n)} \|_F \nonumber \\
		 \ge & \nu - \bar{\sigma}_n \ge 0.6 \, \nu \;\; \mbox{ for } \;\; n \ge N_4 \nonumber \\
		 > & \gamma_n \, .
		\label{naeq8397} 
\end{align}
Thus, ${\cal E}_0 \subseteq \hat{\cal E}$. Now consider the set complements ${\cal E}_0^c$ and $\hat{\cal E}^c$. For the edge $\{ q,\ell \} \in {\cal E}_0^c$, $\| {\bm \Omega}_0^{(q \ell M_n)} \|_F = 0$. For $n \ge N_4$, w.h.p.\ we have
\begin{align}  
		   \| \hat{\bm \Omega}^{(q \ell M_n)} \|_F \le & 
			 \| {\bm \Omega}_0^{(q \ell M_n)} \|_F + \| \hat{\bm \Omega}^{(q \ell M_n)} - {\bm \Omega}_0^{(q \ell M_n)} \|_F \nonumber \\
		 \le & 0 + \bar{\sigma}_n \le 0.4 \, \nu <  \gamma_n \, ,
		\label{naeq8399} 
\end{align}
implying that $\{ q,\ell \} \in \hat{\cal E}_0^c$. Thus, ${\cal E}_0^c \subseteq \hat{\cal E}^c$, hence $\hat{\cal E} \subseteq {\cal E}_0$, establishing $\hat{\cal E} = {\cal E}_0$.
 $\quad \blacksquare$

\bibliographystyle{unsrt}

\begin{thebibliography}{00}


\bibitem{Whittaker1990} J.\ Whittaker, {\em Graphical Models in Applied Multivariate Statistics}. New York: Wiley, 1990.

\bibitem{Lauritzen1996} S.L.\ Lauritzen, {\em Graphical models}. Oxford, UK: Oxford Univ.\ Press, 1996.
 
\bibitem{Buhlmann2011} P.\ B\"{u}hlmann and S.\ van de Geer, {\em Statistics for High-Dimensional data}. Berlin: Springer, 2011.


\bibitem{Meinshausen2006} N.\ Meinshausen and P.\ B\"{u}hlmann, ``High-dimensional graphs and
variable selection with the Lasso,'' {\em Ann.\ Statist.}, vol.\ 34, no.\ 3, pp.\
1436-1462, 2006.


\bibitem{Banerjee2008} O.\ Banerjee, L.E.\ Ghaoui and A.\ d'Aspremont, ``Model selection through
sparse maximum likelihood estimation for multivariate Gaussian or binary
data,'' {\em J.\ Mach.\ Learn.\ Res.}, vol.\ 9, pp.\ 485-516, 2008.

\bibitem{Dahlhaus2000} R.\ Dahlhaus, ``Graphical interaction models for multivariate time series,'' 
 {\em Metrika}, vol.\ 51, pp.\ 157-172, 2000.

\bibitem{Jung2015a} A.\ Jung, G.\ Hannak and N.\ Goertz,
  ``Graphical LASSO based model selection for time series,'' {\em IEEE Signal Process.\ Lett.}, vol.\ 22, no.\ 10, pp.\ 1781-1785, Oct.\ 2015.
	
\bibitem{Tugnait18c} J.K.\ Tugnait, ``Graphical modeling of high-dimensional time series,'' in {\em Proc.\ 52nd Asilomar Conf.\ Signals, Systems, Computers}, Pacific Grove, CA, Oct.\ 29 - Oct.\ 31, 2018, pp.\ 840-844.
	
\bibitem{Tugnait20} J.K.\ Tugnait, ``Consistency of sparse-group lasso graphical model selection for time series,'' in {\em Proc.\ 54th Asilomar Conf.\ Signals, Systems, Computers}, Pacific Grove, CA, Nov.\ 1-4, 2020, pp.\ 589-593.

\bibitem{Tugnait22c} J.K.\ Tugnait, ``On sparse high-dimensional graphical model learning for dependent time series,'' {\em Signal Process.}, vol.\ 197, pp.\ 1-18, Aug. 2022, Article 108539. (Also {\it  arXiv:2111.07897v3} [eess.SP], 4 Jun 2024.) 

\bibitem{Wainwright2019} M.J.\ Wainwright, {\em High-Dimensional Statistics: A Non-Asymptotic Viewpoint}. Cambridge, UK: Cambridge Univ.\ Press, 2019.

\bibitem{Tugnait22d} J.K.\ Tugnait, ``Sparse-group log-sum penalized graphical model learning for time series,'' in {\it Proc.\ IEEE Int.\ Conf.\ Acoust.\ Speech Signal Process.\ (ICASSP), 2022}, pp.\ 5822-5826, Singapore, May 22-27, 2022.

\bibitem{Candes2008} E.J.\ Cand\`{e}s, M.B.\ Wakin and S.P.\ Boyd, ``Enhancing sparsity by reweighted $\ell_1$ minimization,'' {\it J.\ Fourier Anal.\ Appl.}, vol.\ 14, pp.\  877-905, 2008.


\bibitem{Avventi2013}  E.\ Avventi, A.\ Lindquist, and B.\ Wahlberg, ``ARMA identification of graphical models,'' {\it IEEE Trans.\ Autom.\ Control}, vol.\ 58, no.\ 5, pp.\ 1167-1178, 2013.


\bibitem{Alpago2018} D.\ Alpago, M.\ Zorzi and A.\ Ferrante, ``Identification of sparse reciprocal graphical models,'' {\it IEEE Control Sys.\ Lett.}, vol.\ 22, no.\ 4, pp.\ 659-664, 2018.

\bibitem{Songsiri2010} J.\ Songsiri and L.\ Vandenberghe, ``Topology selection in graphical models
of autoregressive processes,'' {\it J.\ Mach.\ Learn.\ Res.}, vol.\ 11, pp.\ 2671-2705, Oct.\ 2010.

\bibitem{You2022}  J.\ You, C.\ Yu, J.\ Sun and J.\ Chen, ``Generalized maximum entropy based identification of graphical ARMA models,'' {\it Automatica}, vol.\ 141, pp.\ 110319, 2022.

\bibitem{Basu2015} S.\ Basu and G.\ Michailidis, ``Regularized estimation in sparse high-dimensional time series models,'' {\it Annals Statistics}, vol.\ 43, no.\ 4, pp.\ 1535-1567, 2015.

\bibitem{Kolar2014} M.\ Kolar, H.\ Liu and E.P.\ Xing, ``Graph estimation from multi-attribute data,'' {\em J.\ Mach.\ Learn.\ Res.}, vol.\ 15, pp.\ 1713-1750, 2014. 

\bibitem{Tugnait21a} J.K.\ Tugnait, ``Sparse-group lasso for graph learning from multi-attribute data,'' {\it IEEE Trans.\ Signal Process.}, vol.\ 69, pp.\ 1771-1786, 2021. (Corrections, vol.\ 69, p.\ 4758, 2021.)

\bibitem{Marjanovic18} G.\ Marjanovic and V.\ Solo, ``Vector $l_0$ sparse conditional independence graphs,'' in {\it Proc.\ IEEE Int.\ Conf.\ Acoust.\ Speech Signal Process.\ (ICASSP), 2018}, pp.\ 2731-2735, 2018.

\bibitem{Yue20} Z.\ Yue, P.\ Sundaram and V.\ Solo, ``Fast block-sparse estimation for vector networks,'' in {\it Proc.\ IEEE Int.\ Conf.\ Acoust.\ Speech Signal Process.\ (ICASSP), 2020}, pp.\ 5505-5509, 2020.

\bibitem{Sundaram20}  P.\ Sundaram, M.\ Luessi, M.\ Bianciardi, S.\ Stufflebeam, M. H\"am\"al\"ainen and V.\ Solo, ``Individual resting-state brain networks enabled by massive multivariate conditional mutual information,'' {\it IEEE Trans.\ Med.\ Imaging}, vol.\ 39, pp. 1957-1966, 2020.

\bibitem{Zhang2017} S.\ Zhang, B.\ Guo, A.\ Dong, J.\ He, Z.\  Xu and S.X.\ Chen, ``Cautionary tales on air-quality improvement in Beijing,'' {\em Proc.\ Royal Soc.\ A}, vol.\ 473, p.\ 20170457, 2017.

\bibitem{Chen2015} W.\ Chen, F.\ Wang, G.\ Xiao, J.\ Wu and S.\ Zhang, ``Air quality of Beijing and impacts of the new ambient air
quality standard,'' {\em Atmosphere}, vol.\ 6, pp.\ 1243-1258, 2015.

\bibitem{Friedman2010a} J.\ Friedman, T.\ Hastie and R.\ Tibshirani, ``A note on the group lasso and a sparse group lasso,'' {\em arXiv:1001.0736v1 [math.ST]}, 5 Jan 2010.
	
\bibitem{Simon2013} N.\ Simon, J.\ Friedman, T.\ Hastie and R.\ Tibshirani, ``A sparse-group lasso,'' {\em J.\ Comput.\ Graphical Statist.}, vol.\ 22, pp.\ 231-245, 2013.

\bibitem{Fan2001} J.\ Fan and R.\ Li, ``Variable selection via nonconcave penalized likelihood and its
oracle properties,'' {\it J.\ Am.\ Statist.\ Assoc.}, vol.\ 96, pp.\ 1348-1360, Dec.\ 2001.


\bibitem{Lam2009} C.\ Lam and J.\ Fan, ``Sparsistency and rates of convergence in large covariance matrix estimation,'' {\em Ann.\ Statist.}, vol.\ 37, no.\ 6B, pp.\ 4254-4278, 2009.

\bibitem{Wolstenholme2015} R.J.\ Wolstenholme and A.T.\ Walden, ``An efficient approach to graphical modeling of time series,'' {\em IEEE Trans.\ Signal Process.}, vol.\ 64, no.\ 12, pp.\ 3266-3276, June 15, 2015.

\bibitem{Luftman2016} D.\ Schneider-Luftman, ``p-Value combiners for graphical modelling of EEG data in the frequency domain,'' {\em J.\ Neuroscience Methods}, vol.\ 271, pp.\ 92-106, 2016.

\bibitem{Matsuda2006} Y.\ Matsuda, ``A test statistic for graphical modelling of multivariate time series,'' {\em Biometrika}, vol.\ 93, no.\ 2, pp.\ 399-409, 2006.

\bibitem{Tugnait19d} J.K.\ Tugnait, ``Edge exclusion tests for graphical model selection: Complex Gaussian vectors and time series,'' {\it IEEE Trans.\ Signal Process.}, vol.\ 67, no.\ 19, pp.\ 5062-5077, Oct.\ 1, 2019.

\bibitem{Songsiri2009} J.\ Songsiri, J.\ Dahl and L.\ Vandenberghe, ``Graphical models of autoregressive processes,'' in Y.\ Eldar and D.\ Palomar (eds.), {\em Convex Optimization in Signal Processing and Communications}, pp.\ 89-116, Cambridge, UK: Cambridge Univ.\ Press, 2009

\bibitem{Eichler2006} M.\ Eichler, ``Graphical modelling of dynamic relationships in multivariate time series,'' in B.\ Schelter, M.\ Winterhalder and J.\ Timmer (eds.), {\em Handbook of time series analysis: Recent theoretical developments and applications}, pp.\ 335-372, New York: Wiley-VCH, 2006.
		
\bibitem{Eichler2012} M.\ Eichler, ``Graphical modelling of multivariate time series,'' {\em Probability Theory and Related Fields}, vol.\ 153, issue 1-2, pp.\ 233-268, June 2012.

\bibitem{Tugnait2022s}  J.K. Tugnait, ``Graph learning from multivariate dependent time series via a multi-attribute formulation,'' in {\it Proc.\ IEEE Int.\ Conf.\ Acoust.\ Speech Signal Process.\ (ICASSP), 2022}, pp. 4508-4512, Singapore, May 22-27, 2022.

\bibitem{Tugnait2021d} J.K.\ Tugnait, ``Sparse-group non-convex penalized multi-attribute graphical model selection,'' in {\it Proc.\ 29th European Signal Process.\ Conf.\ (EUSIPCO 2021)}, pp.\ 1850-1854, Dublin, Ireland, Aug.\ 23-27, 2021. 
	
\bibitem{Zou2008} H.\ Zou and R.\ Li, ``One-step sparse estimates in nonconcave penalized likelihood models,'' {\em Ann.\ Statist.}, vol.\ 36, no.\ 4, pp.\ 1509-1533, 2008.

\bibitem{Tugnait2024} J.K.\ Tugnait, ``Conditional independence graph estimation from multi-attribute dependent time series,'' in {\it Proc.\ IEEE International Workshop on Machine Learning for Signal Processing (MLSP-2024)}, pp.\ 1-6, London, UK, Sept.\ 22-25, 2024.

\bibitem{Loh2017}  P.-L.\ Loh and M.J.\ Wainwright, ``Support recovery without incoherence: A case for nonconvex
regularization,'' {\it Ann.\ Statist.}, vol.\ 45, pp.\ 2455-2482, 2017.
	
\bibitem{Brillinger} D.R.\ Brillinger, {\em Time Series: Data Analysis and Theory}, Expanded 
   edition. New York: McGraw Hill, 1981.
	
	\bibitem{Yuan2007} M.\ Yuan and Y.\ Lin, ``Model selection and estimation in regression with grouped variables,'' {\em J.\ Roy.\ Statist.\ Soc., Ser.\ B (Methodol.)}, vol.\ 68, no.\ 1, pp.\ 49-67, 2006.

\bibitem{Boyd2010} S.\ Boyd, N.\ Parikh, E.\ Chu, B.\ Peleato and J.\ Eckstein, ``Distributed optimization and statistical learning via the alternating direction method of multipliers,'' {\em Found.\ Trends Mach.\ Learn.}, vol.\ 3, no.\ 1, pp.\ 1-122, 2010.

\bibitem{Zhao2022} B.\ Zhao, Y.S.\ Wang and M.\ Kolar, ``FuDGE: A method to estimate a functional differential graph in a high-dimensional setting,'' {\it J.\ Mach.\ Learn.\ Res.}, vol.\ 23, pp.\ 1-82, 2022.

\bibitem{Ravikumar2011} P.\ Ravikumar, M.J.\ Wainwright, G.\ Raskutti and B.\ Yu, ``High-dimensional covariance estimation by minimizing $\ell_1$-penalized log-determinant divergence,'' {\em Electron.\ J.\ Statist.\ }, vol.\ 5, pp.\ 935-980, 2011.

	
\bibitem{Tsay2010} R.S.\ Tsay, {\em Analysis of Financial Time Series}, 3rd Ed., Hoboken, NJ: 
   John Wiley, 2010.

\bibitem{Rothman2008} A.J.\ Rothman, P.J.\ Bickel, E.\ Levina and J.\ Zhu,  ``Sparse permutation invariant covariance
estimation,'' {\em Electron.\ J.\ Statist.\ }, vol.\ 2, pp.\ 494-515, 2008.
	
\bibitem{Schreier10} P.J.\ Schreier and L.L.\ Scharf, {\em Statistical Signal Processing of Complex-Valued Data}, Cambridge, UK: 
   Cambridge Univ.\ Press, 2010.

\bibitem{Boyd2004} S.\ Boyd and L.\ Vandenberghe, {\em Convex Optimization}, Cambridge, UK: 
   Cambridge Univ.\ Press, 2004.

\bibitem{Liu2010} H.\ Liu, K.\ Roeder and L.\ Wasserman, ``Stability approach to regularization selection (StARS) for high dimensional graphical models,'' in {\it Proc.\ NIPS 2010}, pp.\ 1432-1440, 2010.

\bibitem{Wang2010} W.\ Wang, M.J.\ Wainwright and K.\ Ramchandran,``Information-theoretic bounds on model selection for Gaussian Markov random fields,'' in {\em Proc.\ IEEE Int.\ Symp.\ Inf.\ Theory}, Austin, TX, USA, June 2010, pp.\  1373-1377. 
	
\bibitem{Hannak2014}  G.\ Hannak, A.\ Jung, and N.\ G\"{o}rtz, ``On the information-theoretic limits of graphical model selection for Gaussian time series,'' in {\it Proc.\ Eur.\ Signal Process.\ Conf.}, Lisbon, Portugal, 2014, pp.\ 516-520.

          
\end{thebibliography}

\end{document}